\documentclass[10pt,journal,compsoc]{IEEEtran}

\usepackage{bibnames}
\usepackage{amssymb}
\usepackage{verbatim}

\usepackage{amsmath}
\usepackage{amsthm}
\usepackage{graphicx}
\usepackage{caption}
\usepackage{subcaption}
 \usepackage[linesnumbered, ruled,vlined]{algorithm2e}
\usepackage{listings}
\usepackage[flushleft]{threeparttable}
\usepackage{multirow}
\usepackage{listings}
\usepackage{cite}
\usepackage{hyperref}
\usepackage{mdframed}
\usepackage{tabularx}
\usepackage{flushend}
\usepackage{makecell}
\usepackage[T1]{fontenc}
\usepackage{booktabs}
\usepackage[switch]{lineno}
\usepackage{balance}

\usepackage[table]{xcolor}\definecolor{light-gray}{gray}{0.92} 
\newenvironment{gtheorem}%
{\begin{mdframed}[backgroundcolor=light-gray,
skipabove=5pt,
skipbelow=0pt,
nobreak=false
]\begin{mdtheorem}{name}{label}}%
{\end{mdtheorem}\end{mdframed}}
\definecolor{ao}{rgb}{0.0, 0.5, 0.0}
\newcommand{\etal}{\textit{et al.}\space}
\newcommand{\tool}{\textsc{Avastra}\xspace}
\newcommand{\dc}{\textit{DeepCollision}\xspace}
\begin{document}

\title{Generating Critical Scenarios for Testing Automated Driving Systems}

\author{Trung-Hieu~Nguyen,
        Truong-Giang~Vuong,
        Hong-Nam Duong,
        Son~Nguyen,
        Hieu~Dinh~Vo, \\
        Toshiaki Aoki,
        and~Thu-Trang Nguyen
\IEEEcompsocitemizethanks{
    \IEEEcompsocthanksitem Trung-Hieu~Nguyen,
        Truong-Giang~Vuong,
        Hong-Nam Duong,
        Son Nguyen,
        ~Hieu~Dinh~Vo,
        and~Thu-Trang~Nguyen are with the Faculty of Information Technology, University of Engineering and Technology, Vietnam National University, Hanoi.\protect\\
         E-mail: {\{21020017, 21020014, 21020024, sonnguyen, hieuvd, trang.nguyen\}@vnu.edu.vn}
    \IEEEcompsocthanksitem  Toshiaki Aoki is  with School of Information Science, Japan Advanced Institute of Science and Technology\protect\\
    E-mail:{toshiaki@jaist.ac.jp}
    \IEEEcompsocthanksitem Corresponding author: Thu-Trang Nguyen. \protect\\
    }
}


\IEEEtitleabstractindextext{%

\begin{abstract}
Autonomous vehicles (AVs) have demonstrated significant potential in revolutionizing transportation, yet ensuring their safety and reliability remains a critical challenge, especially when exposed to dynamic and unpredictable environments. Real-world testing of an Autonomous Driving System (ADS) is both expensive and risky, making simulation-based testing a preferred approach. 
In this paper, we propose \tool, a Reinforcement Learning (RL)-based approach to generate realistic critical scenarios for testing ADSs in simulation environments.
To capture the complexity of driving scenarios, \tool comprehensively represents the environment by both the \textit{internal states} of an ADS under-test (e.g., the status of the ADS's core components, speed, or acceleration) and the \textit{external states} of the surrounding factors in the simulation environment (e.g., weather, traffic flow, or road condition). 
\tool trains the RL agent to effectively configure the simulation environment that places the AV in dangerous situations and potentially leads it to collisions. 
We introduce a diverse set of actions that allows the RL agent to systematically configure both \textit{environmental conditions} and \textit{traffic participants}.
Additionally, based on established safety requirements, we enforce heuristic constraints to ensure the realism and relevance of the generated test scenarios.
\tool is evaluated on two popular simulation maps with four different road configurations. Our results show \tool's ability to outperform the state-of-the-art approach by generating 30\% to 115\% more collision scenarios. Compared to the baseline based on Random Search, \tool achieves up to 275\% better performance. These results highlight the effectiveness of \tool in enhancing the safety testing of AVs through realistic comprehensive critical scenario generation.
\end{abstract}


}

\maketitle

\IEEEpeerreviewmaketitle

\section{Introduction}

\textit{Autonomous vehicles (AVs)} have shown great potential in transforming modern transportation. The global market for self-driving vehicles is estimated to reach 2.3 trillion USD by 2030~\cite{market-size-forecast}. 
Many automakers, including Tesla, General Motors, and Volkswagen, etc., are actively developing vehicles that are capable of driving autonomously. 
In 2017, Waymo launched the fully autonomous ride-hailing services after extensive testing on the public roads in Arizona~\cite{waymo2017}. Tesla released its beta package of full self-driving software in at least 60.000 vehicles in 2022~\cite{tesla2022}. 

An AV, also known as an \textit{ego vehicle}, is controlled by an \textit{Autonomous Driving System (ADS)} that integrates multiple sensors and software to monitor the driving environment and operate the vehicle.
Ensuring AVs can self-drive without collisions is crucial for the widespread acceptance on real-world roads.
While testing AVs in real-world operating environments is necessary to evaluate their performance in diverse, dynamic, and unpredictable scenarios, it is both expensive and risky. Real-road testing often requires extensive resources, including controlled test tracks, highly instrumented vehicles, and adherence to regulatory standards, making it a costly process. Moreover, exposing AVs, particularly in their early stages of development, to public roads presents significant safety risks, especially when encountering rare or dangerous edge cases that could result in collisions or system failures.
Therefore, it is essential to thoroughly test ADSs in high-fidelity simulators, e.g., LGSVL~\cite{lgsvl} or CARLA~\cite{carla} before deploying.

In simulation-based testing, the \textit{operating environment} is often simulated in various ways to expose the potential failures or weaknesses in ADSs. The operating environment includes a wide range of parameters that present the surrounding space in which ADSs operate. For example, these parameters may include environmental conditions like time of day, weather conditions, road structures, and surrounding obstacles such as other vehicles or pedestrians. Several existing testing approaches~\cite{mosat, av-fuzzer, deepcollision} aim to generate critical scenarios by systematically searching for suitable configurations of these parameters that could challenge the ADS's operation.

Search-based techniques have been widely adopt in various approaches~\cite{mosat,av-fuzzer, blattner2024diversity}.  
For example, MOSAT~\cite{mosat} adopts a multi-objective genetic algorithm 
to search for the atomic or sequence of maneuvers of the other non-player character (NPC) vehicles, which could challenge the ADS and lead it to unsafe scenarios. 
In another study, AV-Fuzzer~\cite{av-fuzzer} perturbs the driving maneuvers of NPC vehicles and employs genetic and fuzzing algorithms to search for scenarios where the AV can dangerously collide with the other vehicles.

Although search-based techniques have succeeded in certain contexts, they face \textit{a major limitation in scalability}, 
due to the vast search space required to model complex driving environments.
Additionally, in real-world scenarios, the safety of AVs is threatened not only by the behaviors of other vehicles but also by numerous other factors, such as pedestrians, weather conditions, and road infrastructure.
However, existing approaches, such as MOSAT~\cite{mosat} and AV-Fuzzer~\cite{av-fuzzer}, focus primarily on the behaviors of NPC vehicles, failing to adequately account for these additional environmental parameters when generating critical scenarios. 
This oversight limits the robustness of their testing and the ability to simulate the full spectrum of potential risks.

To address the challenge of the large and dynamically changing search space of environment parameters, Lu~\etal~\cite{deepcollision} proposed \dc, a Reinforcement Learning (RL)-based approach that interacts with the environment to discover configurations that potentially lead the AV  to crashes. 
While \dc demonstrates effectiveness in generating critical scenarios as the environment evolves, its representation of environment states has limitations. 
Specifically, this approach \textit{does not fully capture the detailed status of the AV and its operating environment}, which could limit the learning ability of the RL agent. Furthermore,  \textit{the realism of the generated test scenarios has not been considered}, potentially resulting in resource inefficiencies when addressing non-critical or unrealistic scenarios.

Building on the promising results of RL in handling the complexity and high dimensionality of the operating environment, we propose \tool, an RL-based approach for generating realistic critical scenarios for testing ADSs in simulation environments.
In \tool, \textit{the environment is comprehensively represented by both the \textbf{internal states of the ADS} (e.g., the status of the ADS's core components, speed, or acceleration) and the \textbf{external states of the surrounding factors} (e.g., weather, traffic flow, or road condition)}.
The RL agent is trained to effectively interact with the simulation environment by taking actions like configuring the weather conditions, introducing pedestrians/vehicles, and other changes that challenge the AV and potentially lead it to collisions. 
In this work, we introduce a diverse set of actions that allows the RL agent to systematically configure both \textit{\textbf{environmental conditions}} and \textit{\textbf{traffic participants}}.
Based on established safety requirements~\cite{jamaframework}, \textit{we incorporate several heuristic constraints} during configuring the environment to ensure the generated scenarios are plausible and relevant for testing.

To evaluate the proposed approach, we conducted several experiments on four roads of two popular maps. The experimental results show that \tool outperforms the baselines in generating critical scenarios. Specifically, the number of collisions generated by \tool is greater than the state-of-the-art approach, \dc, from \textbf{30\% to 115\%}. Additionally, compared to Random Search, \tool obtains up to \textbf{275\%} better in performance.

In summary, the main contributions of this paper are:
\begin{itemize}
    \item \tool, an RL-based approach to generate critical scenarios for testing ADSs
    \item Comprehensively represent the environment in the RL problem by both the internal states of the ADS and external states of the surrounding factors
    \item A diverse set of actions with incorporated constraints for realistically configuring the environmental conditions and traffic participants
    \item An extensive experimental evaluation showing the performance of the proposed approach
\end{itemize}

The implementation of our approaches and the experimental results can be found on our website~\cite{website}.

\textbf{Paper structure:}
Sec.~\ref{sec:problem_formulation} formulates the problem of generating critical scenarios for testing ADSs. Sec.~\ref{sec:solution_formulation} models our solution as an RL problem. The detail of approach is introduced in Sec.~\ref{sec:approach}.
 The experimental setup and results are presented in Sec.~\ref{sec:evaluation_methodology} and Sec.~\ref{sec:results}. 
 Sec.~\ref{sec:related_work} reviews the literature and related studies.
 Finally, Sec.~\ref{sec:conclusion} concludes the paper.

\section{Problem Formulation}
\label{sec:problem_formulation}
A \textit{driving scenario} (or simply \textit{scenario}) is a temporal development of the environment in which the AV operates~\cite{deepcollision, calo2020generating, zhang2021finding, ulbrich2015defining}. Each scenario includes a map describing the road structure, an \text{origin}-\text{destination} pair that defines the starting point and the intended goal of the AV, various environmental conditions (e.g., weather, time of day, and the presence of surrounding static and dynamic obstacles). 
In our work, a \textit{critical scenario} is defined as one that places the AV at risk of colliding with surrounding obstacles.

Let $\mathbb{E} = \{e_1, ..., e_n\}$ represent the set of environmental parameters that are used to define a scenario, where $e_i \in \mathbb{E}$ corresponds to a specific element of the environment (e.g., weather, time of day, pedestrians, or other vehicles).
Given a map $\textbf{M}$ and a pair of origin and destination positions ($P_{org}$, $P_{dest}$) within the map, the set of specific values assigned to the parameters in $\mathbb{E}$ at time step $t$ during the operation of the AV from $P_{org}$ to $P_{dest}$ is denoted by $E_t$. 

The scenario over a time horizon $T$ is therefore described by a sequence of tuples $(\textbf{M}, \mathbb{E}, P_{org}, P_{dest}, \{E_t\}_{t=1}^T)$, where $E_t$ represents the environment observed at time step $t$ during the operation of the AV from $P_{org}$ to $P_{dest}$ on $\textbf{M}$.
The objective of this study is to find $E^*$, the optimal sequence of configurations for the environmental parameters $\{E_t\}_{t=1}^T$, that maximizes the likelihood of collisions during the AV's operation over time. By identifying the most challenging environmental conditions, we aim to push the limits of the ADS and uncover its potential failure points.

During the scenario, the ADS continuously observes the environment and adjusts its own states to navigate safely.
Let $\mathbb{V} = \{v_1, ..., v_m\}$ represent the set of properties that describe the ADS's internal states.
%
Each $v_i \in \mathbb{V}$ captures a specific aspect of the vehicle's behavior (e.g., speed, steering angle, or ADS component statuses). 
In other words, the vehicle's internal state $V_t$ at time step $t$ is dynamically adjusted by the AV in response to the observed environment $E_t$.

\section{Critical Scenario Generation as Reinforcement Learning}
\label{sec:solution_formulation}

In order to find optimal values $E^*$ for the parameters in $\mathbb{E}$, \tool formulates this problem as a Reinforcement Learning (RL) task, represented by a 6-tuple $\langle S, A, T, R, \gamma, \pi \rangle$.

Specifically, $S$ is the \textit{state space} that describes the environment with which the agent interacts.
\tool holistically represents the environment by both the external states of the surrounding factors and the internal states of the ADS's components.
To effectively capture meaningful and relevant information for decision-making within the scenario, as well as to reduce the dimension of state space, 
not all external and internal parameters are directly included in the RL state. Some parameters are transformed or abstracted. 
At each time $t$, the \textit{state} $s_t \in S$ is composed of:
\begin{itemize}
    \item \textbf{Transformed external environmental states from $E_t$}: These represent the transformed/aggregated/selected values of environmental parameters $e_i \in \mathbb{E}$, like time of day, weather, road condition, and the presence of surrounding vehicles or pedestrians.

    \item \textbf{Transformed internal vehicle states from $V_t$}: These represent the selected, transformed, or aggregated properties $v_i \in \mathbb{V}$ that describe the ego vehicle's internal state, such as speed or acceleration. 
\end{itemize}
The state at time step $t$ is represented as $s_t = \tau(\{E_t, V_t\})$, where $\tau$ is the transformation procedure which selects/transforms/aggregates the parameters' values in $E_t$ and ones in $V_t$. This transformation process helps to not only reduce the state space but also allows the RL agent to effectively observe both the external environment and the ego vehicle's internal statuses. 
%
%
The detailed transformation ($\tau$) for these parameters will be discussed in Sec.~\ref{sec:states}.

Each \textit{action} $a_t \in A$ corresponds to setting a specific value for one or more parameters in $\mathbb{E}$, such as modifying weather conditions or adding pedestrians/NPC vehicles. 
After executing action $a_t$, the state of the environment transitions to a new state $s_{t+1}$ according to the \textit{transition function} $T(s_{t+1}|s_t, a_t)$. 
Based on the newly observed state $s_{t+1}$, the agent then receives an immediate \textit{reward} $r_t$ from the \textit{reward function} $R(s_t, a_t)$, which quantifies the effectiveness of taking action $a_t$ in the given state $s_t$. 
The agent receives a higher reward when its selected action leads to conditions that increase the likelihood of collisions for the AV. The goal is to maximize the possibility of generating critical scenarios where the AV is at risk of collisions with other obstacles.
A \textit{discount factor} $\gamma \in [0, 1]$ is used to prioritize immediate rewards over future rewards, encouraging the agent to take actions that lead to quick increases in the likelihood of collisions while still considering the long-term impact of actions.
The probability that the agent will select an action $a_t$ in the given state $s_t$ is decided by the \textit{policy} $\pi: S \times A \to [0, 1]$. The objective of \tool is to find a policy $\pi$ that maximizes the cumulative reward associated with the likelihood of collisions occurring in the scenarios.

\subsection{State Representation}
\label{sec:states}

To enable the RL agent to comprehensively understand the driving scenarios, \tool holistically represents the environment by both the \textit{internal} and \textit{external states}. The \textbf{\textit{internal states}} refer to the ADS’s internal properties and system status, while the \textit{\textbf{external states}} describe the external factors of the environment in which the ADS operates.

Both internal and external states are critical for assessing the performance and safety of the ADS. External factors, such as weather conditions or the presence of dynamic obstacles, can affect the ADS's perception and decision-making, thereby increasing the risk of collisions. Similarly, internal factors, such as localization, speed, and system diagnostics, are also crucial. For example, incorrect localization could lead the AV to make dangerous movements and result in collisions. Thus, an accurate and holistic representation of both the internal and external states is necessary to fully capture the operating conditions of the AV and to evaluate its ability to respond to critical situations.

\subsubsection{External States}
\label{sec:external_states}

External states refer to the environmental factors that could influence the ADS's operation. 
In this work, we design the external states following the perception disturbance factors and characteristics which defined by the ``JAMA Automated Driving Safety Evaluation Framework''~\cite{jamaframework}.
Specifically, we focus on four main external factors, including \textit{weather conditions}, \textit{time of day},  \textit{traffic conditions},
and \textit{road conditions}. 
In particular, \textit{weather conditions} refer to the presence and severity of weather phenomena, including rain, fog, and wetness. \textit{Time of day} is categorized by periods like morning, noon, and night. \textit{Traffic conditions} involve the statuses of the nearest ahead traffic light and traffic flow represented by the number of surrounding obstacles and the information (e.g., speed, distance to the ADS) of the nearest vehicle, which poses the greatest risk to the ADS. 
\textit{Road conditions} describe the type and quality of the road structure, such as one-way road, or intersection, etc., each representing unique challenges to the ADS. The details of these external states are shown in Table~\ref{tab:external_state}.



\begin{table}[]\centering
\caption{Parameter defining \textit{external states} in \tool}
\label{tab:external_state}
\scriptsize
\resizebox{\columnwidth}{!}{%
\begin{tabular}{l|l|l|p{2.5cm}}\toprule
\textbf{No.} & \textbf{Type}    & \textbf{State}             & \textbf{Description} \\\midrule
1           & Weather condition & \textit{weather}           &  - The weather phenomena (rain, fog, and wetness) and their severity levels \\\midrule
2           & Time of day       & \textit{timeOfDay}         &  - Periods in a day (morning, noon, or night) \\\midrule
3           & Traffic light     & \textit{trafficLight}      &  - The traffic light's color which is ahead and nearest to the ego vehicle \\\midrule
4           & Traffic flow      & \textit{numObs}            & - The number of the surrounding obstacles \\
5           &                   & \textit{distToNearestObs}  & - The distance to the nearest obstacle \\
6           &                   & \textit{speedNearestObs}   & - The speed of the nearest obstacle \\ \midrule
7           &  Road condition   & \textit{roadCondition}          & - The type and quality of the road structure (e.g., one-way road or intersection)\\
\bottomrule
\end{tabular}
}
\end{table}
\subsubsection{Internal States}
Internal states refer to the internal properties within the ADS that reflect the operational status of its core components. An ADS typically consists of five main components: \textit{Localization}, \textit{Perception}, \textit{Prediction}, \textit{Planning}, and \textit{Control}~\cite{first-look}. 
For fully autonomous driving, the ADS must first \textit{localize} its position on a high-definition (HD) map. Next, it gathers essential information about the surrounding environment through multiple \textit{perception} sensors and equipment. 
Based on the perception information, the ADS \textit{predicts} the potential trajectories of the nearby pedestrians and vehicles. 
The \textit{planning} module then computes an appropriate trajectory for the ADS to follow.
Finally, the \textit{control} module adjusts the vehicle’s throttle, brake, and steering to execute the planned trajectory. 
Each of these modules plays a critical role in the ADS's safe operation, as errors in any of them can lead to dangerous situations. 
Therefore, in addition to the external states, it is essential to capture these internal states to enable the RL agent to effectively learn about conditions that could lead the AV to potential risks.

\textbf{Localization.} 
An ADS often relies on multiple Global Navigation Satellite System (GNSS) sensors to localize its positions in an HD map. 
%
%
Safe operation requires localization accuracy within centimeter-level~\cite{wan2018robust}, as maneuvers like lane changes and turns depend on precise positioning. 
Inaccurate localization increases the risk of dangerous maneuvers and accidents.

%
\tool captures localization inaccuracies as part of the internal states, measured by the deviation between the actual position (provided by the simulator) and the localized position (determined by the ADS's localization module). 
Formally, let $(x_{local}, y_{local})$ represent the coordinates of the AV's position as obtained from the vehicle's localization module, and $(x_{real}, y_{real})$ represent its actual coordinates on the map. The differences between these positions are quantified by the distance and angle deviations, which measured by the following equations:

\begin{equation}
\label{eq:localization_distance}
 distDeviation = \sqrt{(x_{local} - x_{real})^2 + (y_{local} - y_{real})^2}
\end{equation}

\begin{equation}
\label{eq:localization_angle}
angleDeviation = \arctan \displaystyle \frac{y_{local} - y_{real}}{x_{local} - x_{real}}
\end{equation}

\textbf{Perception.} 
To collect essential information about the surrounding environment, the ADS employs multiple sensors and equipment such as cameras, radars, and LiDAR. By combining data from these sources, the perception module recognizes and predicts the location and movement of the nearby obstacles. 
%
Perception errors, such as misidentifying objects or incorrectly assessing their distances, can lead to inappropriate responses or incorrect decisions, increasing the risk of collisions.

\tool encodes perception-related internal states by measuring the differences between the perceived information and the actual data of the nearby objects.
These differences are measured in terms of position (Equation~\ref{eq:perception_position}), direction of movement (Equation~\ref{eq:perception_direction}), size (Equation~\ref{eq:perception_size}), and velocity (Equation~\ref{eq:perception_velocity}). Note that there could be more than one nearby object, and each object could have an individual set of these attributes. Without loss of generality, we consider the maximum difference as the representative value for each attribute. The reason is that the maximum difference refers to the most significant error in the ADS's perception regarding the corresponding attribute. 

Let $N$ be the number of nearby objects, $(x_{per}^i, y_{per}^i)$ be the perceived coordinates of the object $i$, and $(x_{real}^i, y_{real}^i)$ be its real ones. The difference in position is calculated as:

\begin{equation}
\label{eq:perception_position}
{
\textit{positionDiff} = \max_{i \in [1, N]} \sqrt{(x_{per}^i - x_{real}^i)^2 + (y_{per}^i - y_{real}^i)^2}
}
\end{equation}

Let $dir_{per}^i$ be the perceived moving direction of the object $i$ and $dir_{real}^i$ be its real moving direction. Equation~\ref{eq:perception_direction} shows the difference in moving direction.

\begin{equation}
\label{eq:perception_direction}
\textit{directionDiff} = \max_{i \in [1, N]} |dir_{per}^i - dir_{real}^i|
\end{equation}

Let $h_{per}^i, w_{per}^i$, and $l_{per}^i$ be the perceived sizes regarding height, weight, and length of the object $i$ and $h_{real}^i, w_{real}^i$, and $l_{real}^i$ be its real values. The difference regarding the objects' sizes is computed as:

{
\scriptsize
\begin{equation}
\label{eq:perception_size}
\textit{sizeDiff} = \max_{i \in [1, N]} \sqrt{(h_{per}^{i} - h_{real}^{i})^2 + (w_{per}^{i} - w_{real}^{i})^2 + (l_{per}^{i} - l_{real}^{i})^2}
\end{equation}
}

Let $vx_{per}^i$ and $vy_{per}^i$ be the perceived velocity of the object $i$, and $vx_{real}^i, vy_{real}^i$ be its real velocity values.
The difference in velocity is calculated as:

{
\begin{equation}
\label{eq:perception_velocity}
\textit{velocityDiff} = \max_{i \in [1, N]} \sqrt{(vx_{per}^i - vx_{real}^i)^2 + (vy_{per}^i - vy_{real}^i)^2}
\end{equation}
}

\textbf{Prediction and Planning.}
The prediction module employs various machine learning/deep learning models to predict the trajectories of the surrounding objects, such as pedestrians or vehicles. 
%
Then, their predicted paths help guide the planning module's decisions. 
The planning module calculates a trajectory for the AV, optimizing it for safety, efficiency, and comfort. 
The trajectory is continuously updated as the vehicle moves and new sensor data is received.

Although prediction and planning are critical, \tool does not explicitly represent them in the internal states for two reasons. 
First, at the time of prediction, the predicted and planned trajectories have not yet been executed, making their evaluation challenging. 
Second, the inputs of these modules come from the localization and perception modules, and their outputs are reflected through the operations of the control module. 
Since the states of localization, perception, and control are encoded, the effects of prediction and planning are indirectly represented.

\textbf{Control.}
The control module is responsible for executing the trajectory planned by the planning module. It employs various control algorithms to ensure the vehicle follows the planned trajectory accurately. These algorithms can be categorized into two main types: Lateral control (steering control) and Longitudinal control (speed and braking control). To represent the internal states of the control module, \tool captures several key values that manage the AV's movement, including throttle, 
steering rate, steering target, acceleration rate, brake percentage, and speed.

Instead of representing module operations through errors as we do with the localization and perception modules, we capture the actual value of control parameters as internal states of the control module. Indeed, localization and perception modules play crucial roles in the AV's behaviors and safety, but these modules only \textit{indirectly} affect how AV moves. They impact the AV's movement by providing information to the other modules like planning and control.
%
%
Any errors in the localization and perception modules could lead to incorrect decisions that result in dangerous situations. Thus, we represent the operations of these modules through their errors to assess their accuracy. Meanwhile, the control module is the final stage in the ADS operation, which \textit{directly} affects AV's movement and safety. 
Capturing the actual values of the control parameters allows us to understand the exact behavior and movement of the AV on the road, which directly contribute to potential safety risks.

The details of internal states are shown in Table~\ref{tab:interal_states}.

\begin{table*}\centering
\caption{Parameter defining \textit{internal states} in \tool}
\label{tab:interal_states}
\begin{tabular}{l|l|l|p{8cm}}\toprule

\textbf{No.} &\textbf{Module} &\textbf{State} &\textbf{Description} \\\midrule
1           & Localization  &\textit{distDeviation}  &- The distance deviation of the located position and the actual position of the AV \\
2           &               &\textit{angleDeviation} &- The angle deviation of the located position and the actual position of the AV \\\midrule
3           & Perception    &\textit{positionDiff}   &- The difference between the perception of the AV and the actual information of the positions of the nearby objects \\
4           &               &\textit{directionDiff}  &- The difference between the perception of the AV and the actual information of the moving directions of the nearby objects \\
5           &               &\textit{sizeDiff}       &- The difference between the perception of the AV and the actual information of the size of the nearby objects \\
6           &               &\textit{velocityDiff}   &- The difference between the perception of the AV and the actual information of the velocities of the nearby objects \\ \midrule
7           & Control       &\textit{throttle}       &- Specify how much fuel is delivered to the AV's engine \\
8           &               &\textit{steerRate}      &- The rate at which the AV's steering angle changes \\
9           &               &\textit{steerTarget}    &- The intended steering angle that the AV aims to achieve \\
10          &               &\textit{accRate}        &- The rate at which the AV is accelerating \\
11          &               &\textit{brakePercentage} &- The amount of braking force applied \\
12          &               &\textit{speed}          &- Driving speed of the AV \\
\bottomrule
\end{tabular}
\end{table*}

\textit{In summary}, at a time step $t$, a state $s_t \in S$ is defined as a state-tuple $\langle w_1, ..., w_{19}\rangle$, where $w_i$, $1 \leq i \leq 19$, is the value of the $i^{th}$ state variable. The first seven state variables are external states (Table~\ref{tab:external_state}), and the remaining 12 variables are internal states (Table~\ref{tab:interal_states}). Together, these variables comprehensively represent both the external environmental factors and the internal operations of the ADS.
\subsection{Action Space}
\label{sec:action}

Based on established safety requirements~\cite{jamaframework}, we design the action space to configure the environmental parameters, aiming to generate critical and realistic scenarios for testing the ADS. In practice, the operating environment could be extremely complicated with potentially infinite parameters. Simulating such complexity in a practical testing setup would be computationally prohibitive. 
However, when testing the ADS in a simulated environment, the number of parameters available for configuration is limited by the simulator in use. 
In this work, we employ the widely-used simulator LGSVL~\cite{lgsvl}, which supports configurable environment parameters across four primary categories: \textit{Weather}, \textit{Time of Day}, \textit{Pedestrians}, and \textit{NPC Vehicles}. 
Each category can be divided into sub-types, allowing for detailed control of the environment. For example, \textit{Weather} includes two sub-types of \textit{Phenomenon} (i.e.,  \textit{Rain}, \textit{Fog}, and \textit{Wetness}) and the corresponding intensity \textit{Level} (i.e., \textit{None}, \textit{Light}, \textit{Moderate}, or \textit{High}).
%
In general, our approach can be extended to other simulators, such as CARLA~\cite{carla} or AirSim~\cite{airsim}, which offer similar capabilities.
To configure the environment, an action involves choosing parameter(s) and specifying specific value(s). For example, setting the weather to heavy rain by assigning \textit{Rain} to the parameter phenomenon and \textit{High} to the parameter level accordingly.
In \tool, we design a diverse set of actions allowing to configure both the \textit{environmental conditions} and \textit{traffic participants}. Environmental conditions include setting weather conditions and time periods. Meanwhile, configuring traffic participants involves introducing NPC vehicles and pedestrians with different behaviors into the operating map. The action groups are summarized in Table~\ref{tab:action_group}, and the possible parameter values are listed in Table~\ref{tab:parameter_value}.

While the flexibility of the simulator allows for many configurable parameters and possible values, a naive approach to action construction could lead to an exponentially large and impractical action space, with many combinations resulting in unrealistic or infeasible scenarios. 
Therefore, to prevent the combinatorial explosion of parameter values and the generation of impractical scenarios, we incorporate several \textit{heuristic constraints} to ensure that the actions reflect real-world scenarios.

\begin{table}\centering
\caption{Group of actions in \tool}
\label{tab:action_group}
\resizebox{\columnwidth}{!}{%
\begin{tabular}{l|l|p{4.7cm}}\toprule \textbf{No.} & \textbf{Group} & \textbf{Specific action} \\\midrule
1 & Time of day & \{$time$\} \\ \midrule
2 & Weather condition & \{$level$\} \{$phenomenon$\} \\ \midrule
3 & NPC drive-ahead &\{$vehicle\_type$\} with  \{$vehicle\_color$\} in \{$distance$\} at \{$speed$\} is driving ahead in the \{$lane$\} lane \\ \midrule
4 & NPC overtake & \{$vehicle\_type$\} with \{$vehicle\_color$\} in \{$distance$\} at \{$speed$\} is overtaking in the \{$lane$\} lane \\ \midrule
5 & NPC drive-opposite &\{$vehicle\_type$\} with \{$vehicle\_color$\} in \{$distance$\} at \{$speed$\} is driving opposite \\ \midrule
6 & NPC cross-road &\{$vehicle\_type$\} with \{$vehicle\_color$\} in \{$distance$\} at \{$speed$\} is crossing road \\ \midrule
7 & NPC lane-change & \{$vehicle\_type$\} with \{$vehicle\_color$\} in \{$distance$\} at \{$speed$\} is changing from the \{$lane$\} lane to \{$lane$\} lane\\ \midrule
8 & Pedestrian cross-road & A pedestrian in \{$distance$\} at \{$speed$\} is crossing road from the \{$lane$\} lane to \{$lane$\} lane  \\
\bottomrule
\end{tabular}
}
\end{table}

\begin{table}\centering
\caption{Values for the parameters in \tool's actions}\label{tab:parameter_value}
\begin{tabular}{l|p{4.7cm}}\toprule
\textbf{Parameter} &\textbf{Values} \\\midrule
$\textit{time}$ &\{Morning, Noon, Night\} \\\midrule
$\textit{level}$ & \{None, Light, Moderate, High\} \\ \midrule
$\textit{phenomenon}$ &\{Rain, Fog, Wetness\} \\\midrule
$\textit{vehicle\_type}$ &\{Jeep, BoxTruck, Sedan, SUV, SchoolBus, Hatchback\} \\\midrule
$\textit{vehicle\_color}$ &\{Pink, Red, Yellow, Black, White, Skyblue\} \\\midrule
$\textit{distance}$ & $d \in \{n | n \in \mathbb{N}^+\}$ \\\midrule
$\textit{speed}$ & $s \in \{n | n \in \mathbb{N}^+\}$  \\\midrule
$\textit{lane}$& \{Current, Left, Right\}\\ 
\bottomrule
\end{tabular}
\end{table}

\textbf{Chronological time constraint.}
The time of day must progress in chronological order. Let $time(t)$ represent the value of the time of day at time point $t$.
For example, if the state $time$ is \textit{Night} at $t$, i.e., $time(t) = \textit{Night}$, it cannot abruptly change to $\textit{Noon}$ in the next time point $t + 1$. Thus, the constraints on the value of the $time$ parameter for each subsequent time point $t + 1$ are as follows: 
$time(t) = \textit{Morning} \implies time(t + 1) \in \{\textit{Morning}, \textit{Noon}\} $,
$time(t) = \textit{Noon} \implies time(t + 1) \in \{\textit{Noon}, \textit{Night}\} $,
and 
$time(t) = \textit{Night} \implies time(t + 1) \in \{\textit{Night}, \textit{Morning}\}$.

\textbf{Traffic Participant Constraints.} 
To test the ADS's responses to nearby traffic participants, \tool gradually introduces NPC vehicles and pedestrians with diverse behaviors into the operating map. 
NPC vehicles have five possible behaviors: $\textit{NPC\_behavior} \in$ \{$\textit{drive-ahead}$, $\textit{overtake}$, $\textit{drive-opposite}$, $\textit{cross-road}$, $\textit{lane-change}$\}. 
For a pedestrian, we consider only road-crossing behavior, $\textit{Pedestrian\_behavior} \in \{\textit{cross-road}\} $, since it poses potential risks to the ADS's performance. In fact, pedestrians can perform other behaviors, such as walking along the sidewalk; however, such behavior is generally safe and rarely impacts the ADS. 

To generate \textit{challenging} and \textit{realistic} scenarios, it is crucial to position the NPC vehicles or pedestrians (\textit{obstacles}) such that their potential trajectories should intersect with that of the ego vehicle to create a risk of collision. However, obstacles should not be introduced too abruptly or closely so that a collision becomes unavoidable regardless of the ADS's capabilities. A realistic scenario gives the ADS a \textit{fair} opportunity to detect and react to potential threats. 
%

%
%

To guarantee that the scenarios are both challenging and realistic, we propose constraints to determine the position of introducing the obstacles, i.e., constraints for the $\textit{distance}$ parameter in the actions related to NPC vehicles and pedestrians. Specifically, the distance from the AV to the position of the generating obstacle is measured based on three factors: distance mode ($\textit{dist\_mode} \in \{\textit{Near}, \textit{Far}\}$), vehicle size ($\textit{size} \in \{\textit{Small}, \textit{Large}\}$), and \textit{speed}. 
These factors significantly influence how much time and space the ADS needs to respond to the obstacles. 
First, the distance mode (\textit{dist\_mode}) defines basic settings for testing the ADS's reactions to the obstacles. Intuitively, the closer an obstacle is, the higher the likelihood of a collision. The farther an obstacle is, the more difficult it is for the detection models. Thus, \tool generates obstacles in both \textit{Near} and \textit{Far} cases to comprehensively assess the ADS's performance. 
Second, vehicle size (\textit{size}) also significantly impacts ADS's vision and reaction. Larger vehicles often occupy more road space and pose a greater threat to the ADS due to their size and limited maneuverability. Therefore, \tool generates both \textit{Small} and \textit{Large} vehicles to test the ADS's reactions.
Third, \textit{speed} represents how fast the ego vehicle and the obstacle approach each other. Hence, the speed values are also leveraged to estimate the initial distance of the obstacle so that a collision could happen.

\begin{equation}
\label{eq:minimum_distance_constraint}
    \small
    \textit{distance} = \alpha*f_{d}(\textit{dist\_mode}) + \beta*f_{v}(\textit{size}) + \gamma*f_s(\textit{speed})
\end{equation}

Without loss of generality, the distance for generating an obstacle is defined by Equation~\ref{eq:minimum_distance_constraint}. In this Equation, $(\alpha, \beta, \gamma)$ are weight coefficients for each factor, which can be adjusted according to their relative importance. In this work, \tool considers them equally important, with $\alpha = \beta = \gamma = 1$. $f_{d}(\textit{dist\_mode})$ defines the base minimum distance, which specifies the setting of \textit{Near} or \textit{Far}. For instance, we could set $f_{d}(Far) = 50m$ and $f_{d}(\textit{Near}) = 12m$. $f_{v}(size)$ specifies the additional distance required according to the vehicle size. Large vehicles require greater distance since they pose a greater challenge to ADS safety. For example, \tool can set $f_{v}(\textit{Small}) = 0m$ and $f_{v}(\textit{Large}) = 3m$. \textit{Large} vehicles include BoxTruck and SchoolBus, while the \textit{Small} vehicles are Jeep, Sedan, SUV, and Hatchback. $f_s(speed)$ determines the extra distance required based on the obstacle's speed. Higher-speed obstacles need more extra distance because they can reach the (estimated) intersection point of their and the ego vehicle's (potential) trajectories more quickly. 
If this distance is not increased accordingly, the obstacle could pass the intersection point before the ego vehicle arrives, reducing the effectiveness of the generated scenario. Therefore, \tool incorporates extra distance adjustments based on the obstacle's speed. The specific values of these functions are empirically determined via various experiments. The detailed distance for generating an obstacle in each action can be found on our website~\cite{website}.

Moreover, the relative position of the NPC vehicle to the ego vehicle's lane (i.e., on the \textit{Current}, \textit{Left}, or \textit{Right} lane) also affects the ADS's operation because it impacts how the ADS perceives, predicts, and plans responses to NPC vehicles. Therefore, in generating NPC vehicles, \tool considers all the three possible values of the $lane$ parameter. 
The $speed$ of an NPC vehicle is randomly set within the range of $[10, 20]$ \textit{m/s}, reflecting the typical speed of cars on the road. Additionally, $speed$ is also guaranteed to be below the speed limitation of the operating road. 
The pedestrians' speed is guaranteed to be no greater than the normal human walking pace, $1.4$ \textit{m/s}.
Furthermore, to enhance the diversity of the generated scenarios, the values of $\textit{vehicle\_type}$ and $\textit{vehicle\_color}$ are randomly selected for each action.

\textbf{Weather pattern constraints.}
Weather phenomena and their intensity levels are interdependent. For example, it would be unrealistic for \textit{Wetness} to be at \textit{High} level without the presence of \textit{Rain} or \textit{Fog}. 
However, precisely modeling the dependencies of weather conditions in real-world scenarios is highly complex. 
To keep our actions manageable, we do not consider weather constraints in this paper. For future work, we plan to collect and analyze real-world datasets to extract realistic weather patterns and use them to construct weather-related actions.

\textit{In summary}, our action space $A$ consists of 45 actions in total, including 13 actions for configuring weather and time of day, 30 actions for generating NPC vehicles with diverse behaviors, and two actions for creating road-crossing pedestrians. 
The detailed value for each parameter of these actions can be found on our website~\cite{website}.

\subsection{Reward Function}
\label{sec:rewards}

After executing action $a_t$, the operating environment transitions from state $s_t$ to the new state $s_{t+1}$, represented by the transition $\langle s_t, a_t, s_{t+1} \rangle$. To provide feedback on the effectiveness of the action $a_t$, \tool rewards the RL agent $r_t$. 
The reward is designed to reflect the preference of the action.
Actions that better align with \tool's objective of generating critical scenarios are preferable and receive higher reward values.
Intuitively, $r_t$ is calculated by a reward function designed based on the \textit{collision probability}~\footnote{ \textit{Collision probability} formula is presented in Sec.~\ref{sec:proc_measurement}.} ($ProC$), where actions that increase this probability are rewarded more. 
Particularly, actions that result in an actual collision will receive the highest reward ($R_{col}$), which is much higher than the rewards of the others. This encourages the agent to prioritize actions that really challenge the ADS.
Specifically, the reward function is formulated as follows:


\begin{equation}
\label{eq:reward_function}
    r_t =  \begin{cases}
        R_{col} , & \text{} ProC = 1.0 \\
        ProC, & \text{} 0.2 < ProC < 1.0 \\
        -1, & \text{otherwise}
    \end{cases}
\end{equation}

In this formula, $ProC$ represents the collision probability, 
and $R_{col}$ is the reward awarded to agent when its chosen action results in an actual collision. If $ProC = 1$, indicating a collision has occurred, the agent receives a reward of $R_{col}$
For $0.2 < ProC < 1$, the selected action places the ego vehicle in a dangerous situation, and receives a reward of $ProC$.
In cases where $ProC \leq 0.2$, indicating safe conditions, the agent receives no reward and is penalized with a reward of $-1$.

\section{RL-based Critical Scenario Generation for testing ADSs}
\label{sec:approach}

\begin{figure*}
    \centering
    \includegraphics[width=\linewidth]{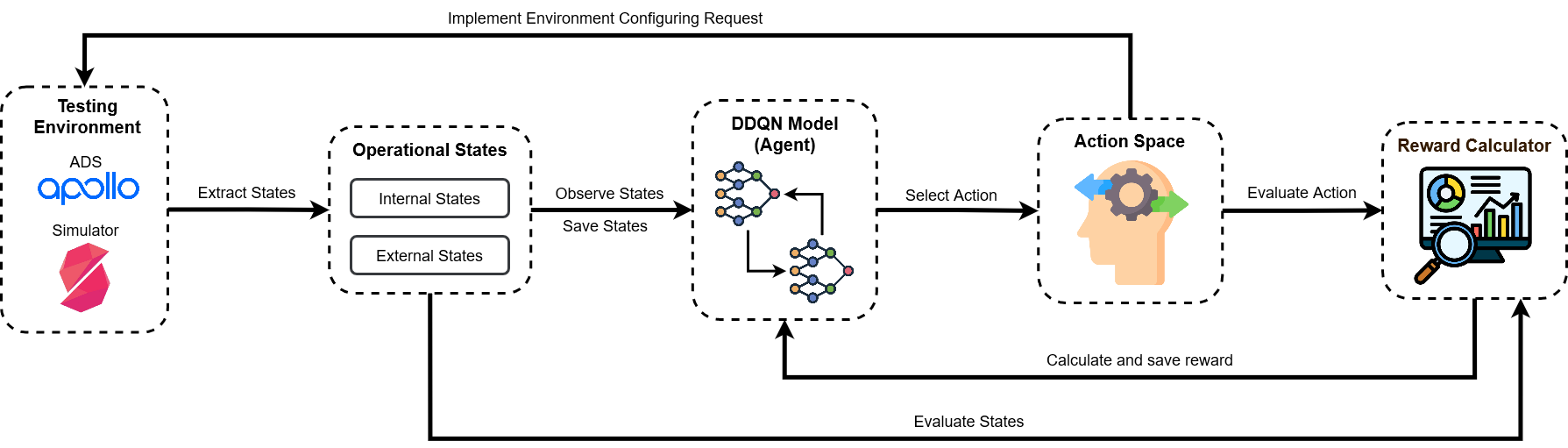}
    \caption{\tool: Approach Overview}
    \label{fig:framework_overview}
\end{figure*}

Figure~\ref{fig:framework_overview} shows the overview of \tool, our approach for generating critical scenarios for testing ADS operations. \tool leverages a Double Deep-Q Network (DDQN) to train an RL agent that can observe both the ADS internal states and surrounding external states to select optimal actions to configure the environment. 
At each time step $t$, based on the observed state $s_t$, the agent selects an action $a_t$. Upon executing $a_t$ and the ADS operates within a fixed observation-time period (OTP), the operating environment is transitioned into a new state $s_{t+1}$. 
The agent then evaluates the effectiveness of the chosen action by calculating a reward based on the \textit{collision probability}, reinforcing actions that lead to more critical scenarios. 
%


Moreover, to effectively train the DDQN model, a \textit{Replay Buffer} with \textit{Prioritized Experience Replay} (PER) algorithm is employed in the training process. Specifically, at each time step $t$, the transition  $\langle s_t, a_t, s_{t+1} \rangle$ along with its corresponding reward $r_t$ is stored into the buffer. When the Replay Buffer reaches its capacity, the transitions and rewards are prioritized using PER, which ensures that high-priority transitions are selected to train the DDQN model.

Each scenario initiates from the origin to a designed destination. However, the scenario may terminate early if the ego vehicle is unable to move forward after 3 to 5 consecutive steps, ensuring that inefficient or deadlock conditions are handled appropriately.

\subsection{Double Deep Q-Network}

For the RL problem, Deep Q-Network (DQN) is a popular algorithm demonstrated as a powerful tool for estimating the Q-values~\cite{fan2020theoretical}. 
However, the inherent stochasticity in the AV's environments, including unpredictable weather condition, traffic statuses, and surrounding objects' behaviors, can lead DQN to overestimate the action's effectiveness due to their reliance on a single network for both action selection and evaluation~\cite{zhang2017weighted}. This results in biased updates that may converge to suboptimal policies.

\begin{figure}
    \centering
    \includegraphics[width=\linewidth]{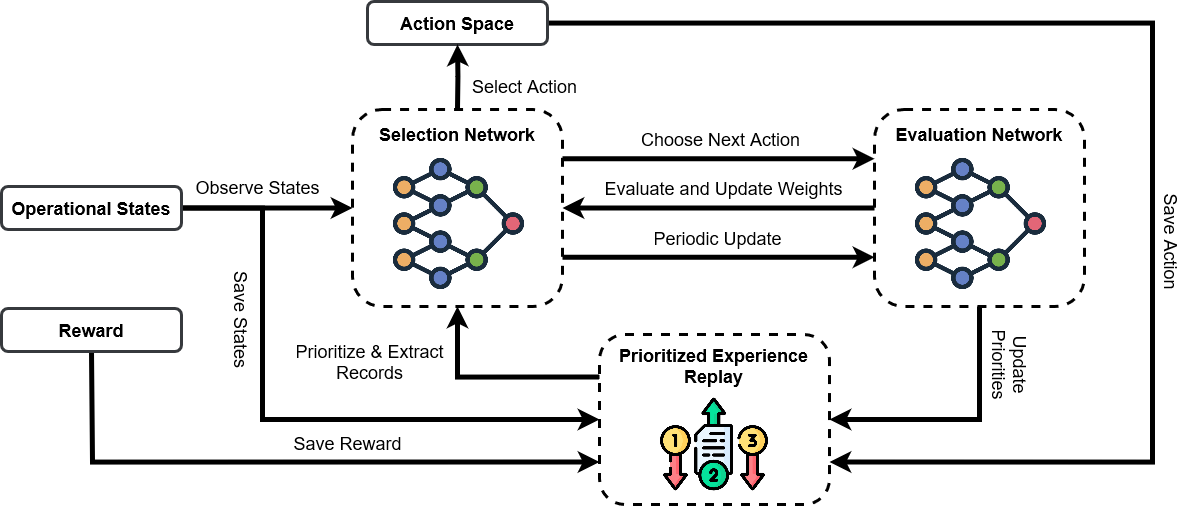}
    \caption{Double Deep-Q Network Architecture in \tool}
    \label{fig:ddqn_model}
\end{figure}

To overcome this, \tool employs DDQN to estimate the expected accumulated reward for each action within a given state, as shown in Figure~\ref{fig:ddqn_model}.
DDQN mitigates the overestimation issue using two separate networks: one for \textit{action selection} and one for \textit{action evaluation}. 
The \textit{action selection} network determines the action with the highest Q-value for a given state, 
%
while the \textit{action evaluation} network evaluates this choice and updates the Q-values based on actual outcomes. 
This separation allows for more accurate evaluation, ensuring that Q-values reflect real-world performance rather than inflated estimates. Periodic updates from the selection network to the evaluation network further stabilize the learning process.

In \tool, both networks share the same architecture, comprising four layers: an input layer, two hidden layers, and an output layer. The input layer processes state vectors that comprehensively represent both the internal and the external states (see Sec.~\ref{sec:states}). The hidden layers extract relevant features, and the output layer estimates Q-values for potential actions. \textit{ReLU} activation functions are used for hidden layers, with \textit{Linear} in the output layer to approximate return for agent.
At each time step $t$, the agent selects an action to configure the environment. 
To avoid premature convergence to local optima, we implement an $\epsilon$-greedy exploration strategy. Initially, $\epsilon$ is set to 1.0, allowing for random exploration as the agent learns the environment dynamics. As training progresses, $\epsilon$ is gradually reduced to 0.1 to balance exploration with exploitation, following the decay schedule from previous studies~\cite{deepcollision, mnih2015human}.

To further improve the training process, \tool employs a Replay Buffer with PER algorithm. 
After executing an action $a_t$, transition $\langle s_t, a_t, s_{t+1} \rangle$ and the corresponding reward $r_t$ are stored into the buffer. The PER algorithm prioritizes these transitions based on 
the accuracy evaluated by the selection network, ensuring that the agent focuses more on poorly evaluated actions. This targeted prioritization reduces bias towards frequent but less impactful transitions, enabling more effective and efficient learning of critical scenario generation.
%

\subsection{Collision Probability Measurement}
\label{sec:proc_measurement}

In this work, we adopt a straightforward approach to calculate the \textit{collision probability}, similar to the method used in prior study~\cite{deepcollision}. The calculation relies on the \textit{safety distance} and the \textit{current distance} between the AV and the other surrounding obstacles in the operating environment. 
Specifically, the safety distance represents the minimum required separation between the AV and other objects to avoid a collision, while the current distance refers to the actual distance at a given moment.
If the current distance is smaller than the safety distance, the AV is in a dangerous situation, and a collision could happen.

\subsubsection{Safety distance}
In practice, the safety distance could be affected by various factors such as road conditions, speed, vehicle performance, driver/ADS reaction time, etc. 
However, these factors cannot be fully accessed in a simulation environment, and it is very complicated to consider them.
To simplify the process, as a simulation-based testing approach, \tool follows the related studies~\cite{deepcollision, cheng2019longitudinal, manjunath2013survey, chen2019vehicle} that model safety distance based on speed and deceleration in both the longitudinal and lateral dimensions. 
Since AVs can react almost instantaneously~\cite{xie2018heterogeneous}, the reaction time is disregarded in the calculation of safety distance in this work.

The longitudinal safety distance (\textit{LoSD}) ensures the AV maintains a safe distance from vehicles ahead and behind in the same lane, while the lateral safety distance (\textit{LaSD}) ensures safe spacing between the AV and vehicles in adjacent lanes. These distances are calculated as follows:

\begin{equation}
\label{eq:loSD}
   \textit{LoSD}(v_f, v_l) = \frac{1}{2} \left(\frac{v_{f}^2}{\alpha_f} - \frac{v_{l}^2}{\alpha_l}\right) + R_{min}
\end{equation}

\begin{equation}
\label{eq:laSD}
   \textit{LaSD}(v_{ego}) = \frac{(v_{ego} \times sin \beta)^2}{\alpha_{ego} \times sin \beta}
\end{equation}

Here, $v_f$, $v_l$, $v_{ego}$ are the velocities of the following, leading, and the ego vehicles, respectively.
Similarly, $\alpha_f$, $\alpha_l$, and $\alpha_{ego}$ denotes their deceleration rates. 
$R_{min}$ refers to the minimum allowable distance between two vehicles, while $\beta$ is the angle between the ego vehicle's direction and the lane where the obstacle is located. 
In this work, we use the default settings from the Berkeley algorithm~\cite{cheng2019longitudinal}, which assigns a deceleration value of $-6 \text{m/s}^2$ for all vehicles and sets $R_{min}$ to 5 meters. 
The values of $v_f$, $v_l$, $v_{ego}$, and $\beta$ can be obtained from the simulator.

\subsubsection{Current distance}

The \textit{current distance} between the ego vehicle and any obstacle is determined using their respective positions.
Let $(x_{ego}, y_{ego}, z_{ego})$ represent the position of the ego vehicle, and $(x_{ob}, y_{ob}, z_{ob})$ be the position of an obstacle. The current distance is then computed as:

\begin{equation}
\label{eq:cd}
 CD = \sqrt{(x_{ego} - x_{ob})^2 + (y_{ego} - y_{ob})^2 + (z_{ego} - z_{ob})^2}
\end{equation}

\subsubsection{Collision probability}
The collision probability is derived from the calculated \textit{Safety Distance} and \textit{Current Distance}, following the existing method~\cite{deepcollision}.
We define \textit{ProC} in two dimensions, longitudinal and lateral, denoted as \textit{loProC} and \textit{laProC}, respectively, to measure the likelihood of a collision with surrounding obstacles. 
Let $N$ be the total number of surrounding obstacles observed at time step $t$; $\textit{loProC}_i$ and $\textit{laProC}_i$ represent the probability that the ego vehicle collides with $i^{th}$ obstacle in two dimensions. These overall longitudinal and lateral collision probabilities of the AV are calculated as follows:

\begin{equation}
\label{eq:loproc}
loProC = \max_{i \in [1, N]} \textit{loProC}_i
\end{equation}
\begin{equation}
\label{eq:laproc}
laProC = \max_{i \in [1, N]} \textit{laProC}_i
\end{equation}
In the formula \ref{eq:loproc} and \ref{eq:laproc}, $loProC_i$ and $laProC_i$ are measured by the following equations: 

\begin{equation}
\label{eq:loproc_i}
loProC_i = \begin{cases}
            \frac{LoSD - CD}{LoSD}, &\text{same lane} \vspace{0.25cm} \\
            0, & \text{different lane}
        \end{cases}
\end{equation}

\begin{equation}
\label{eq:laproc_i}
laProC_i = \begin{cases}
            \frac{LaSD - CD}{LaSD}, &\text{different lane} \vspace{0.25cm} \\
            0, & \text{same lane}
        \end{cases}
\end{equation}
Here, $loProC_i$ and $laProC_i$ range from 0.0 to 1.0, where 1.0 indicates a collision with the $i^{th}$ obstacle, and 0.0 indicates no risk of collision.
Finally, the overall collision probability (\textit{ProC}) is determined by combining both the maximum and minimum values of \textit{loProC} and \textit{laProC}:
\begin{equation}
\label{eq:proc}
\textit{ProC} =  \textit{ProC}_{max} + (1 - \textit{ProC}_{max}) \times \textit{ProC}_{min}
\end{equation}

In Equation~\ref{eq:proc}, $\textit{ProC}_{max} = \max (\textit{loProC}, \textit{laProC})$ and $\textit{ProC}_{min} = \min (\textit{loProC}, \textit{laProC})$.
The value of \textit{ProC} falls into the range of $[\textit{ProC}_{max}, 1]$, whereas $ProC = 1.0$ indicates a collision.

\section{Evaluation methodology}
\label{sec:evaluation_methodology}
\subsection{Research questions}
To evaluate the proposed approach, we aim to address the following research questions:
\begin{itemize}
    \item \textbf{RQ1. Performance Comparison:} How effective is \tool in generating critical scenarios for testing the ADS? How does it compare to the state-of-the-art methods?

    \item \textbf{RQ2. Component Analysis:} How do different components of RL framework such as state space and action space impact \tool's performance in generating critical scenarios? 

    \item \textbf{RQ3. Parameter Analysis:}
    How do different parameters such as $OTP$, $R_{col}$, and $\epsilon$ contribute to the overall results of \tool?

    \item \textbf{RQ4. Sensitivity Analysis:} How does the variety of training environments impact \tool's ability to generate critical scenarios for testing an ADS on a fixed testing road?

\end{itemize}
\subsection{Experimental Procedure and Metrics}

\subsubsection{System, Simulator, and Maps}

\textbf{Subject System:} We adopt Baidu's Apollo\footnote{https://github.com/ApolloAuto/apollo} (version 7.0) as the subject system under test. Apollo is an industrial-grade open-source autonomous driving platform known for its flexible architecture, which supports both software-in-the-loop (SIL) and hardware-in-the-loop (HIL) testing. Widely used in existing research~\cite{deepcollision, mosat, av-fuzzer}, Apollo 7.0 provides a robust foundation for evaluating ADS.

\noindent\textbf{Simulator:} We use the LGSVL Simulator~\cite{lgsvl}, an open-source, Unity-based simulator designed for autonomous driving scenarios. LGSVL offers high-fidelity simulation capabilities, allowing for complex and realistic environments where the autonomous system can be rigorously tested. Its integration with Apollo further ensures consistency between the simulation environment and the ADS under test.

\begin{figure*}
    \centering
    \begin{subfigure}{.6\linewidth}
        \begin{subfigure}{.475\linewidth}
            \centering
            \includegraphics[width=\linewidth]{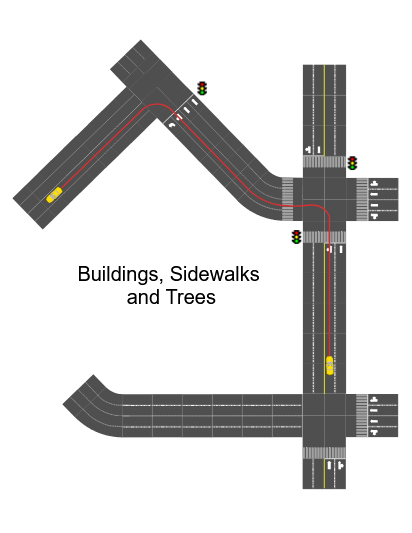}
            \caption{Road 1: \textit{L-Shaped Residential Junction}}
            \label{fig:state_road1}
        \end{subfigure}%
        \hspace{0.5cm}
        \begin{subfigure}{.475\linewidth}
            \centering
            \includegraphics[width=\linewidth]{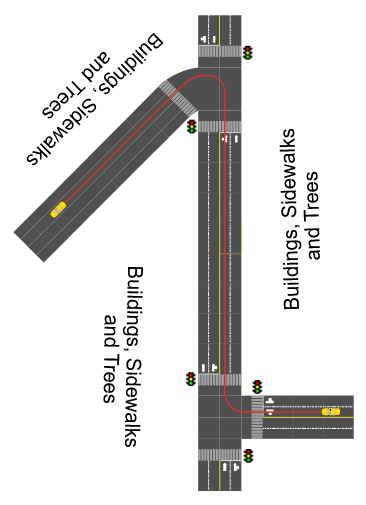}
            \caption{Road 2: \textit{Complex Corner Intersection}}
            \label{fig:state_road1}
        \end{subfigure}%
        \vspace{0.4cm}
        \\ 
        \begin{subfigure}{\linewidth}
            \centering
            \includegraphics[width=\linewidth]{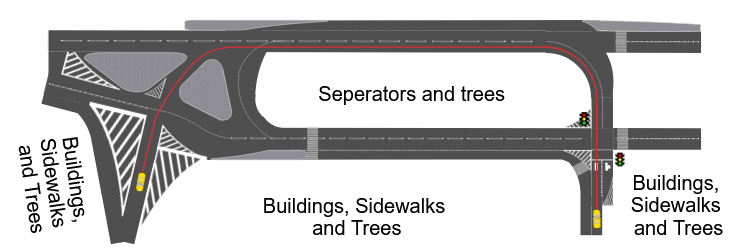}
            \caption{Road 4: \textit{Curved Boulevard}}
            \label{fig:state_road1}
        \end{subfigure}%
    \end{subfigure}
    \hspace{0.5cm}
    \begin{subfigure}{.3\linewidth}
        \centering
        \includegraphics[width=1.1\linewidth]{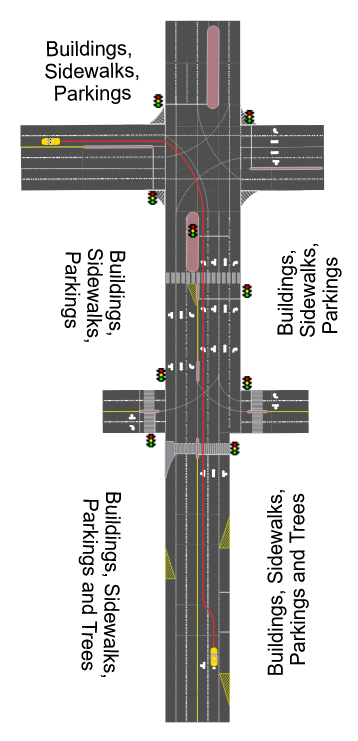}
        \caption{Road 3: \textit{Multi-Lane Urban Crossroad}}
        \label{fig:state_road3}
    \end{subfigure}%
    \caption{Representation of experimental roads in \tool. }
    \label{fig:exp_road}
\end{figure*}

\noindent\textbf{Experimental Maps:} To evaluate the proposed method, we selected the \textbf{San Francisco} map, which is widely used by the existing studies in critical scenario generation~\cite{deepcollision, mosat}. Additionally, we also included another map, i.e., \textbf{Tartu}, for evaluation to evaluate \tool and the other approaches in varied road configurations. 
Each map features distinct characteristics that enhance the diversity of the experiments.
Specifically, we assess two road configurations for each map, making a total of four roads. 
The roads evaluated feature a variety of complex elements, such as traffic lights, intersections, cross-streets, sharp turns, and one-way or two-way roads, as shown in Figure~\ref{fig:exp_road}. These diverse and potentially hazardous road segments provide a thorough test bed for evaluating the robustness of our proposed method in different driving scenarios.
This selection ensures that our evaluations cover a broad range of driving conditions, increasing the reliability and generalizability of our findings.


\subsubsection{Experimental Procedure}

\textbf{RQ1. Performance Comparison}.
To evaluate the effectiveness of \tool, we compare its performance against two baselines: \textit{\dc} and \textit{Random Search}.

\begin{itemize}
    \item \textit{\textbf{Random Search}}: A widely used baseline for evaluating search methods, which selects actions randomly within the action space. It serves as a reference to measure the capability of more sophisticated methods in finding critical scenarios for testing ADS.

    \item \textit{\textbf{\dc}}: The state-of-the-art approach that uses an RL agent trained through continuously interacting with the environment. The goal of the agent is to identify configurations likely to lead the AV to collisions.

\end{itemize}

\textbf{RQ2. Component Analysis.} This experiment studies how different components of the RL framework, such as state and action spaces impact \tool's performance.

\textit{State Analysis}. To understand how different states affect \tool's performance, we consider two types of states: {Internal State} and {External State}. We conduct experiments on two variants: $\tool_{I}$ with only \textit{Internal State}  and $\tool_{E}$ with only \textit{External State}, then compared their results with \tool incorporated  \textit{full state} configuration. 

\textit{Action Analysis.} 
%
%
To assess the \textit{influence of action space}, we 
conduct an ablation study on three groups of actions, including environmental conditions (No. 1-2), NPC vehicles (No. 3-7), and pedestrians (No. 8) in Table~\ref{tab:action_group}.
By comparing the outcomes of \tool with different sets of actions, we would like to highlight the importance of all the groups of actions in generating critical scenarios.
Moreover, we also evaluate the \textit{impact of our heuristic constraints} on the overall results of \tool by comparing the performance of \tool with and without constraints.

\textbf{RQ3. Parameter Analysis.} This experiment examines how different parameters such as $OTP$, $R_{col}$, and $\epsilon$ contribute to the overall results of \tool. We gradually change each parameter's values to analyze how each specific value impacts the quality of generated scenarios. 

\textbf{RQ4. Sensitivity Analysis}.  we evaluate \tool's performance on a fixed testing road (Road 4) using three different training settings: \textit{Cross-Road} Training (training on Roads 1, 2, and 3), \textit{Same-Road} Training (training on Road 4 alone), and \textit{Comprehensive-Road} Training (training on all four roads). For each setting, we measure the effectiveness of \tool in generating critical scenarios. This allows us to assess how training on diverse or identical environments affects \tool's capability in generating challenging and realistic test scenarios on the target road.

For each experimental setting, we train the models over 750 episodes, selecting the versions with the highest or most stable average returns for evaluation.  For each step, after applying a selected action, the ADS operates within an default interval $OTP = 6s$.
During testing, each model is evaluated across 15 episodes, using five different random seeds to ensure robust results. The results are averaged to provide a clear comparison of how each component contributes to \tool's effectiveness in discovering critical scenarios.

\subsubsection{Metrics}
\label{sec:metrics}
We adopted  \textit{\#Collisions}
and \textit{Time to Collision (TTC)}, which are widely used to evaluate the performance of the critical scenario generation methods~\cite{deepcollision,mosat, av-fuzzer}.

\textbf{\#\textit{Collisions}} is the number of collisions between the ego vehicle and the surrounding obstacles in the generated scenarios. The higher the number of collisions observed, the better the approach. 
A collision occurs when the distance between the ego vehicle and an obstacle is less than or equal to zero, i.e., $ProC = 1.0$. 

However, incorrect measurements of actual collisions may arise in certain cases.
For example, after an initial collision, the ego vehicle may continue moving forward, causing repeated collisions with the same obstacle at the same position as previous one. 
In such cases, a collision could be reported multiple times, resulting in an overestimation of the performance of the approaches. 
To ensure accurate performance evaluation, we follow these conventions for counting collisions:
\begin{itemize}
    \item \textbf{Counting only the first collision in repeated sequences}: If the ego vehicle continues to collide with the same obstacle at the same location, only the first collision is counted to prevent overestimation.

    \item \textbf{Excluding subjective collisions}: These are unavoidable collisions caused by environmental configurations beyond the handling capability of the ADS. Our focus is on testing ADS limitations, so we exclude scenarios where the collisions are beyond ADS control:
        \begin{itemize}
            \item \textbf{Sudden obstacles in the safety zone}: If an object, like a pedestrian or vehicle, suddenly appears within the ADS safety zone, it will likely result in an unavoidable collision regardless of the ADS's capabilities. Therefore, such cases
            are considered unrealistic and excluded in the scope of this study.

            \item \textbf{Side or rear-end collision caused by NPC vehicles/pedestrians}: Collisions where pedestrians or NPC vehicles  (intentionally) collide with the AV from the AV's side or rear are also excluded, as such accidents are unavoidable for the AV and 
             beyond the control of the ADS.
        
        \end{itemize}

\end{itemize}

\textbf{\textit{Time to Collision (TTC)}} measures the time to the first collision occurrence in a scenario. A lower value of TTC means that the approach effectively configures the environment to quickly expose the failures of the ADS. The smaller \textit{TTC}, the better the approach.





\section{Experimental results}
\label{sec:results}
\subsection{Answer RQ1: Performance Comparison}

\begin{table}\centering
\caption{Performance comparison}
\label{tab:performance_comparison}
\resizebox{\columnwidth}{!}{%
\begin{tabular}{l|l|l|rr}\toprule
\textbf{Map} &\textbf{Road} &\textbf{Approach} &\textbf{\#Collisions} &\textbf{TTC}\\\midrule

\multirow{6}{*}{San Fran.} 
&\multirow{3}{*}{Road 1} 
    &Random Search  &3.2                &30.2 \\
&   &\dc            &7.6                &26.6 \\
&   &\tool          &\textbf{12.0}      &30.9\\

\cmidrule{2-5}

&\multirow{3}{*}{Road 2} 
    &Random Search  &5.4                &43.3 \\
&   &\dc            &12.6               &27.3 \\
&   &\tool          &\textbf{18.9}      &15.8 \\

\midrule

\multirow{6}{*}{Tartu} 
&\multirow{3}{*}{Road 3} 
    &Random Search  &5.2                &39.0 \\
&   &\dc            &5.4                &7.1 \\
&   &\tool          &\textbf{11.6}      &29.4 \\

\cmidrule{2-5}

&\multirow{3}{*}{Road 4} 
    &Random Search  &7.0                &32.8 \\
&   &\dc            &10.4               &20.9 \\
&   &\tool          &\textbf{13.4}      &32.6 \\
\bottomrule
\end{tabular}
}
\end{table}

To address RQ1, we evaluate the performance of \tool against \dc and Random Search using two key metrics: the number of collisions (\#\textit{\textbf{Collisions}}) and Time to Collision (\textit{\textbf{TTC}}) across four distinct road configurations in San Francisco and Tartu. Each road exhibits different complexities with different layouts and infrastructures.
The results are summarized in Table~\ref{tab:performance_comparison}.

As seen, \tool consistently outperforms the other methods across all road configurations, generating a significantly higher number of collisions. Specifically, on Road 1 (\textit{L-Shaped Residential Junction}), \tool produces 58\% more collisions compared to \dc and about 4 times more collisions than Random Search. Similarly, on Road 2 (\textit{Complex Corner Intersection}), \tool increases the number of collisions by 50\% over \dc and by a substantial 3.5 times compared to Random Search. This trend continues for Road 3 (\textit{Multi-Lane Urban Crossroad}) and Road 4 (\textit{Curved Boulevard}), where \tool yields higher collision rates by 115\% and 30\% compared to \dc, respectively.
Regarding TTC, \tool performs better by generating collision scenarios faster in certain roads, but there are instances where its performance lags behind \dc. 
This suggests that while \tool excels at generating more collision scenarios, it sometimes takes longer to trigger a collision event in simpler or less dynamic environments.

The performance of the ADS is significantly affected by the road complexity. In particular, among the scenarios generated by all the approaches, the ADS encountered the highest number of collisions on Road 2 and the fewest collisions on Road 3. For instance, with scenarios generated by \tool, there were 18.9 collisions on Road 2, while 11.6 collisions on Road 3. Similarly, the figures for scenarios generated by \dc on Road 2 and Road 3 were 12.6 and 5.4 collisions, respectively.
Indeed, Road 2 includes a sharp turn coupled with building proximity, which not only restricts the ADS's perception but also demands precise and quick adjustments in steering, speed, and positioning. 
This complex road layout significantly increases the likelihood of collisions. Meanwhile, Road 3 is a wide-open road with clear separators and fewer tight turns, allowing the ADS to navigate more smoothly and with fewer distractions, thereby reducing the chances of dangerous situations. These results highlight the ADS's limitations on roads with tight intersections and complex layouts.

\begin{figure}
    \centering
    \begin{subfigure}{\columnwidth}
        \begin{subfigure}{.45\columnwidth}
            \centering
            \includegraphics[width=0.6\linewidth]{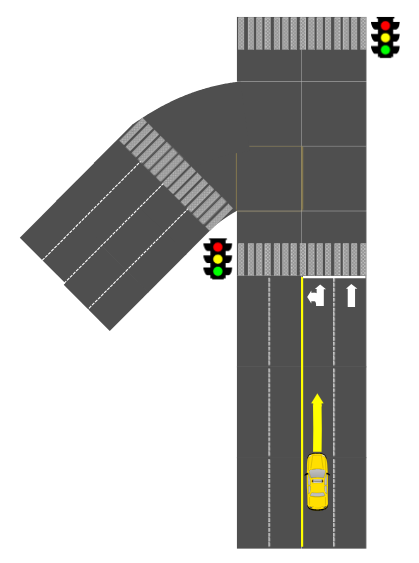}

             \caption{Step 1: \textit{The AV is approaching a  junction. \tool is expected to leverage this traffic condition to create a challenging scenario.\\}}
            \label{fig:bad-step1}
        \end{subfigure}%
        \hspace{0.5cm}
        \begin{subfigure}{.45\columnwidth}
            \centering
            \includegraphics[width=0.6\linewidth]{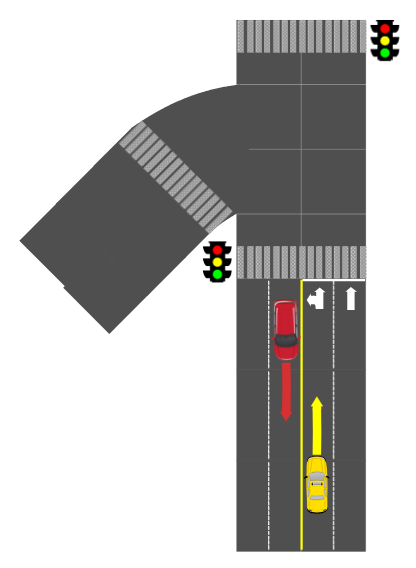}
            \caption{Step 2: \textit{\tool introduces an NPC vehicle at the opposite direction in the adjacent lane. However, the median makes the scenario to be less challenging.}}

            \label{fig:bad-step2}
        \end{subfigure}%
    \end{subfigure}

 \caption{Ineffective scenario generated by \tool on Road2}
    \label{fig:bad-case}
\end{figure}

Analyzing the generated scenarios, we found several cases where \tool demonstrates the limitation due to its lack of consideration for the lane structure.
For example, as illustrated in Figure~\ref{fig:bad-step1}, the AV is approaching a junction and intends to turn to reach its destination.
To challenge the operation of the AV, \tool introduces a red NPC vehicle traveling in the opposite direction in the adjacent lane (Figure~\ref{fig:bad-step2}),
which could pose a collision risk during the AV's lane change or turn.
However, there is a median that physically separates the lanes, preventing the AV from entering the NPC vehicle's lane and versa vise. Thus, the risk of head-on collisions is significantly reduced in this case. Without incorporating the lane structures such as land boundaries or medians, \tool overestimates the risk associated with this NPC vehicle placement. As a result, this scenario is generally ineffective in testing the AV's responses to safety-critical situations.

%


    
   

\begin{figure}
    \centering
    \begin{subfigure}{\columnwidth}
        \begin{subfigure}{.45\columnwidth}
            \centering
            \includegraphics[width=\linewidth]{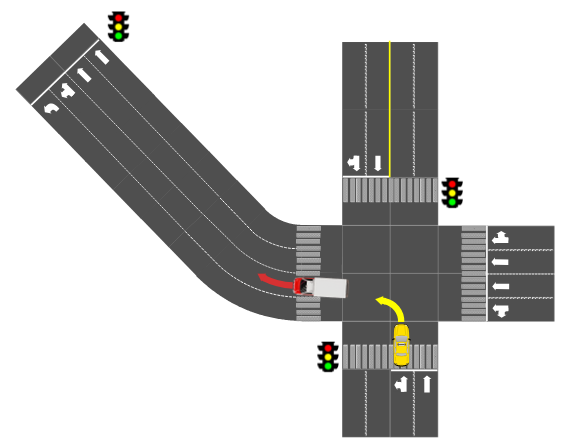}
            \caption{Step 1: \textit{The AV is entering the junction where there is red box truck moving in the same direction.}}
            \label{fig:interesting-case-step1}
        \end{subfigure}%
        \hspace{0.5cm}
        \begin{subfigure}{.45\columnwidth}
            \centering
            \includegraphics[width=\linewidth]{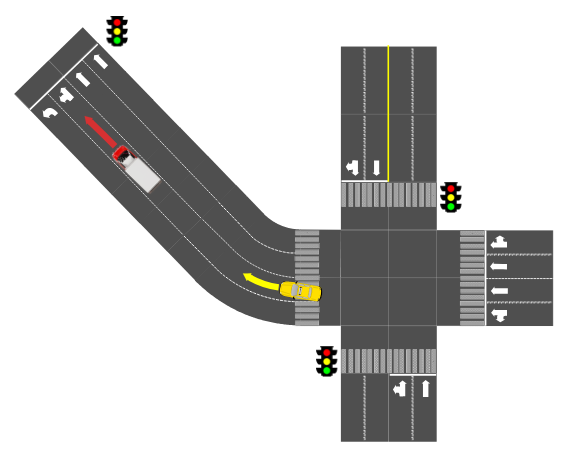}
            \caption{Step 2: \textit{The box truck is moving toward the stop sign, and the AV is following closely at a high speed.
}}
            \label{fig:interesting-case-step2}
        \end{subfigure}%
    \end{subfigure}

\vspace{0.5cm}

    \begin{subfigure}{\columnwidth}
        \begin{subfigure}{.45\columnwidth}
            \centering
            \includegraphics[width=\linewidth]{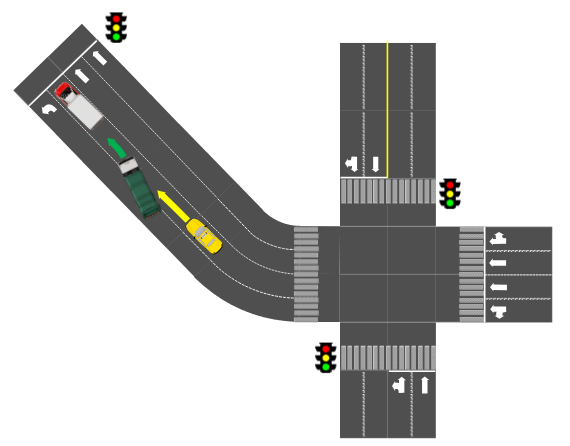}
            \caption{Step 3: \textit{\tool introduces a another green box truck on the left lane. Then, the green box truck changes lane and merges into the AV's lane. }}
            \label{fig:interesting-case-step3}
        \end{subfigure}%
        \hspace{0.5cm}
        \begin{subfigure}{.45\columnwidth}
            \centering
            \includegraphics[width=\linewidth]{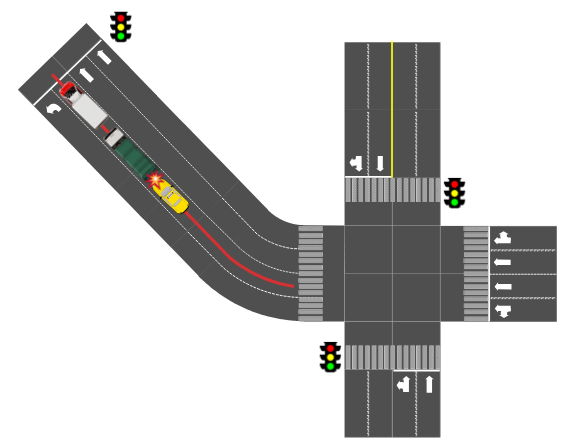}
            \caption{Step 4: \textit{
            Both red and green box trucks are stopping due the red light. However, the AV fails to make proper braking decision and collides with the green box truck.}}
            \label{fig:interesting-case-step4}
        \end{subfigure}%
    \end{subfigure}

    \caption{Critical scenario generated by \tool on Road 1}
    \label{fig:interesting-case}
\end{figure}
%


Figure~\ref{fig:interesting-case} shows an interesting critical scenario where \tool effectively utilizes both the internal states of the AV and the external states of the surrounding factors to create a dangerous situation. First, to challenge the AV's operation when it enters the junction, \tool introduces a red box truck (Figure~\ref{fig:interesting-case-step1}). After entering the junction, the AV continues to follow the box truck to the stop sign (Figure~\ref{fig:interesting-case-step2}). By considering the external states, \tool can capture the traffic light and traffic flow to assess the surrounding environment. As well as incorporating the internal states, \tool is aware that the AV is driving at high speed. Next, \tool introduces a new green box truck (Figure~\ref{fig:interesting-case-step3}), which then merges into the same lane as the AV and the red truck box. Due to a red light, all the vehicles must stop. However, the ADS fails to recognize the presence and the stop of the green box truck. As a result, the AV cannot adjust its speed in time, leading to a collision with the green box truck (Figure~\ref{fig:interesting-case-step4}). This complex scenario tests the AV's ability to handle sudden lane changes and stops, pushing it to adapt quickly under realistic conditions.

These examples reveal both the strengths and weaknesses of \tool. Addressing limitations like lane structure consideration could further enhance \tool's capability, ensuring that it consistently generates realistic, challenging scenarios that thoroughly test the ADS's operation.

\begin{gtheorem}
\textbf{Answer to RQ1}: \tool consistently outperforms the baseline approaches, \dc and Random Search, in generating critical scenarios across all four road configurations. The number of collisions produced by \tool exceeds those of the baseline methods, with improvements ranging from 30\% to 275\%. However, \tool sometimes requires more time to trigger collisions compared to \dc, as reflected by a higher TTC observed in certain roads.
\end{gtheorem}

\subsection{Answer RQ2. Component Analysis}
\subsubsection{State Analysis}
\begin{figure}
    \centering
     \begin{subfigure}{\linewidth}
        \centering
        \includegraphics[width=\linewidth]{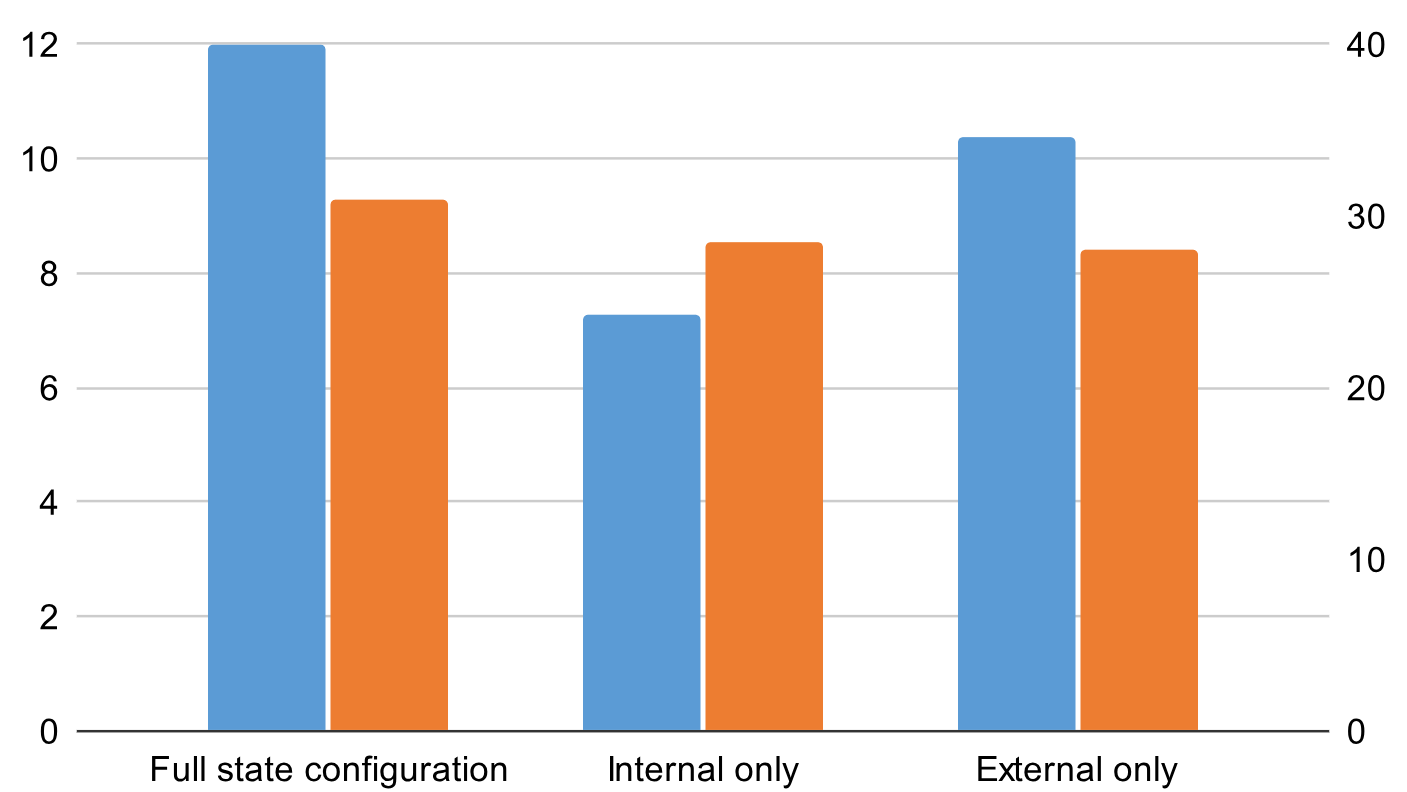}
        \caption{Road 1}
        \label{fig:state_road2}
    \end{subfigure}%
    
    \begin{subfigure}{\linewidth}
        \centering
        \includegraphics[width=\linewidth]{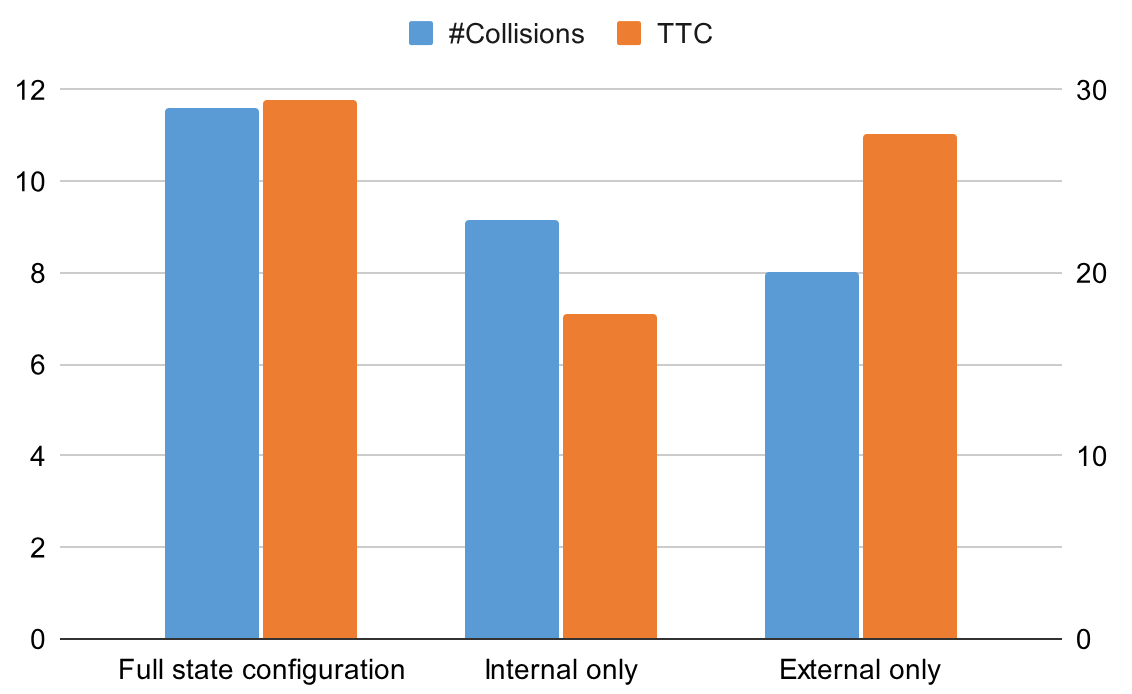}
        \caption{Road 3}
        \label{fig:state_road1}
    \end{subfigure}%

     \caption{State analysis on Road 1 and Road 3 (\textit{\#Collisions}: Left axis; \textit{TTC}: Right axis)}
    \label{fig:state_analysis}
\end{figure}

Figure~\ref{fig:state_analysis} shows that \tool performs significantly better on both Road 1 (\textit{L-Shaped Residential Junction}) in the San Francisco map and Road 3 (\textit{Complex Corner Intersection}) in the Tartu map when using full state space. Utilizing the full state space helps to improve \tool's performance by $15$\% to $65$\% compared to the two versions $\tool_{I}$ (using only internal) and $\tool_{E}$ (using only external states).  
These results indicate that leveraging both the \textit{internal} and \textit{external} states significantly enhances \tool's stability and effectiveness, as it could gain a comprehensive understanding of the extrinsic environment and the intrinsic operations of the ADS.

The results of the individual state variants, $\tool_{I}$ and $\tool_{E}$, varies significantly across the two roads. 
On Road 1, $\tool_{E}$ achieves better performance than $\tool_{I}$. By using only external states, $\tool_{E}$ obtains 41\% more collisions than $\tool_{I}$ which use only internal states on this road. Indeed,
the curved layout of Road 1 with multiple turns puts considerable external challenges on the ADS's performance. To follow the road and navigate these curves accurately, the ADS needs to effectively monitor and analyze the external elements such as road conditions, traffic lights, and traffic flow to make precise movements. 
This makes the external states more relevant for generating critical scenarios on this road.

Meanwhile, $\tool_{I}$ outperforms $\tool_{E}$  on Road 3, suggesting that internal states play a more critical role in this setting. For instance, by considering only external states, 
$\tool_{E}$ recorded eight collisions in the generated scenarios, while with only internal states, the number of collisions in $\tool_{I}$'s generated scenarios increased by 23\%. 
On simple and wide-open road layouts such as Road 3, fewer external challenges exist to  mask the visibility of the ADS. 
In such roads, the collisions are primarily driven by
the ineffectiveness and inaccuracies of the ADS's internal components. Therefore, on Road 3, only focusing on internal states can enable \tool to generate critical scenarios more effectively than relying on only external states.

These variabilities suggest that relying solely on either internal or external states makes \tool sensitive to the specific features of each road, leading to instability in performance.
These findings reinforce the importance of an integrated approach in which \tool utilizes both internal and external states to maintain consistency and effectively handle varying road geometries and traffic dynamics.


\subsubsection{Action Analysis}

\begin{table}[]\centering
\caption{Performance of generating critical scenarios using difference configurations of action-space}
\label{tab:performance_action_space}
\resizebox{\columnwidth}{!}{%
\begin{tabular}{l|rr}\toprule
\textbf{Action space} &\textbf{\#Collisions} &\textbf{TTC}\\\midrule
NPC Vehicles + Weather\&Time &14.4 &24.4  \\
Weather\&Time + Pedestrians  &21.0 &14.4  \\
Pedestrians + NPC Vehicles &19.0 &24.6  \\
NPC Vehicles + Pedestrians + Weather\&Time &20.2 &29.0 \\ 
\bottomrule
\end{tabular}
}
\end{table}


\textbf{Impact of action-space configurations.} Table~\ref{tab:performance_action_space} shows 
the performance of \tool in generating critical scenarios across various action-space configurations.

The configuration with only Pedestrians and NPC vehicle actions results in 6\% fewer collisions than when Weather\&Time actions are also involved. 
This is likely because, without the additional unpredictability of environmental changes, the ADS may perform better at identifying and managing interactions with pedestrians and NPC vehicles, leading to a lower collision rate.
In contrast, complex environmental conditions could reduce the ADS's perception capability. Thus, changing the weather and time conditions increases the likelihood of collisions.

The ADS demonstrates a high sensitivity to pedestrians. By the configuration of only Weather\&Time and Pedestrian actions, \tool obtained the highest number of collisions (21 collisions) in the generated scenarios with the shortest \textit{TTC} (14.4s). 
In contrast, excluding pedestrians led to a 30\% decrease in collisions. This result indicates that the presence of pedestrians poses a significant challenge for the ADS.

While enabling  Weather\&Time and Pedestrian actions yields the highest number of collisions, using the full action space in \tool is still recommended. 
The full action space yields a collision count close to the highest result and, importantly, allows \tool to generate a broader range of scenarios, including collisions involving both vehicles and pedestrians. This diversity helps reveal more types of weaknesses within the ADS, making it a valuable tool for comprehensive safety evaluation.


\begin{table}
\centering
\caption{Impact of  the heuristic constraints  on the performance of \tool}
\label{tab:heuristic_constraints}
\resizebox{\columnwidth}{!}{%
\begin{tabular}{l|rrrr}\toprule
&\textbf{\#Collisions} &\textbf{\#Subj. Coll.} &\textbf{TTC} \\\midrule
\textit{With constraints}        &18.9   &2.2    &15.78 \\
\textit{Without constraints}     &9.0    &4.4    &17.45 \\
\bottomrule
\end{tabular}
}
\end{table}

\textbf{Impact of heuristic constraints.} In this experiment, we evaluate the impact of the heuristic constraints on the performance of \tool. To this end, we built two variants of \tool: one that incorporates the constraints into the actions used for configuring the environment (Sec.~\ref{sec:action}) (\textit{With constraints}), and another without these constraints (\textit{Without constraints}). In the latter variant, the values of the corresponding parameters (i.e., $distance$)  of the actions are randomly selected within their valid ranges. 

As shown in Table~\ref{tab:heuristic_constraints}, incorporating heuristic constraints significantly \textit{improves the generation of challenging scenarios}, as indicated by higher collisions and shorter \textit{TTC}. In particular, \tool with constraints doubles the number of collisions compared to the variant without constraints (18.9 vs. 9.0 collisions). Additionally, \textit{TTC} decreases from 17.45 seconds to 15.78 seconds when the constraints are applied. 
This improvement indicates that our heuristic constraints make environment configuration actions more effective and efficient at generating scenarios that trigger collisions.

Moreover, the constraints also \textit{enhance the realism of the scenarios} by reducing the number of subjective collisions (\#Subj. Coll.). Specifically, subjective collisions occur beyond the controls of the ADS (Sec.~\ref{sec:metrics}) and are less valuable for testing ADS limitations. With the heuristic constraints, the number of subjective collisions is halved, dropping from 4.4 to 2.2. 
This reduction highlights how the constraints help avoid unrealistic scenarios, thereby improving the quality and realism of the generated scenarios.

In practice, without the heuristic constraints, the NPC vehicles/pedestrians are introduced into the scenario at random positions. This randomness often results in two problematic cases: NPC vehicles/pedestrians appearing either too far or too close to the ADS. 
The obstacles too far away fail to sufficiently challenge the ADS's operation, leading to less effective scenarios with fewer collisions.
Meanwhile, the obstacles that suddenly appear too close to the ADS often cause unavoidable collisions as the ADS lacks sufficient time to respond appropriately (higher subjective collisions). However, such abrupt collisions are unrealistic, as NPC vehicles or pedestrians would not ``teleport'' into the ADS's path in real-world traffic but would instead gradually enter its field of view.
Overall, using heuristic constraints ensures more meaningful testing scenarios that better reflect real-world conditions, thereby enabling a more robust evaluation of ADS performance.

\begin{gtheorem}
\textbf{Answer to RQ2}: Leveraging the full state and action spaces enables \tool to achieve optimal performance. Additionally, incorporating constraints during configuring the environment enhances the relevance and the realism of the generated scenarios.
\end{gtheorem}

\subsection{Answer RQ3: Parameter Analysis}

\subsubsection{$OTP$ Analysis}

\begin{figure}
    \centering
    \includegraphics[width=\linewidth]{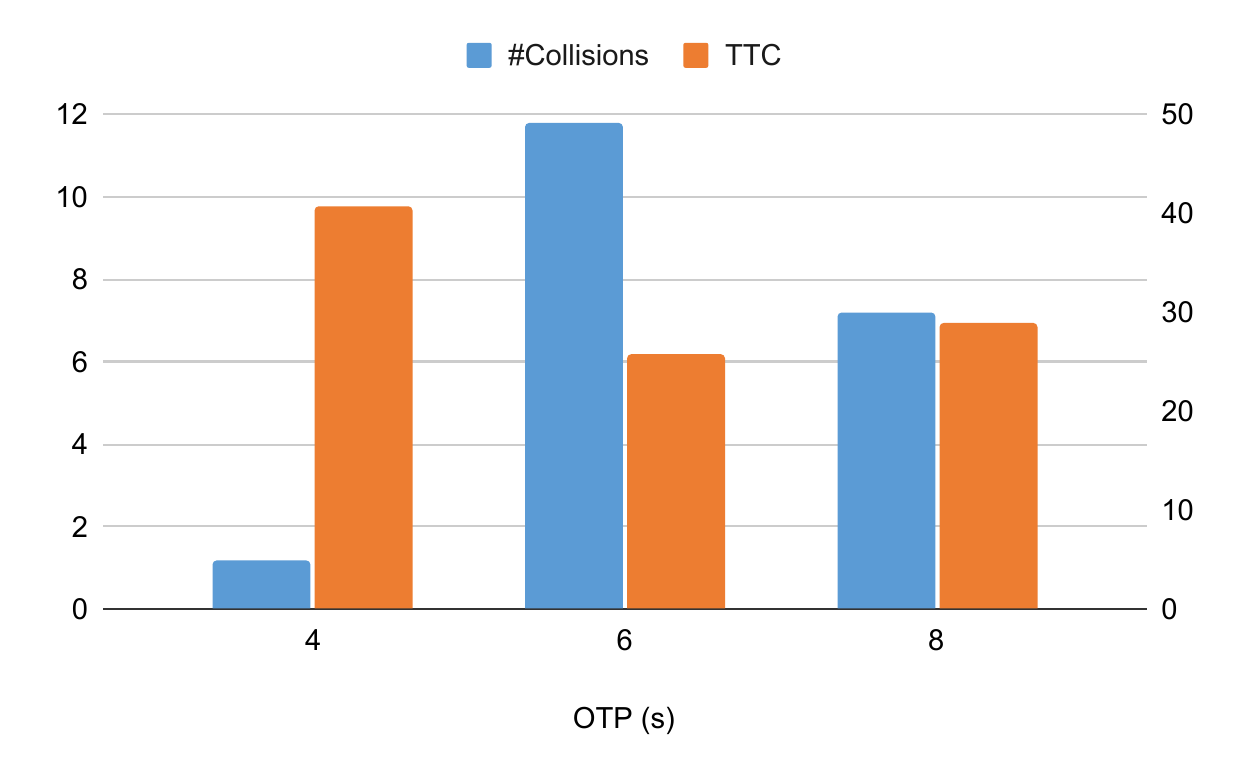}
    \caption{Impact of $OTP$ values on \tool's performance (\textit{\#Collisions}: Left axis; \textit{TTC}: Right axis)}
    \label{fig:otp_analysis}
\end{figure}

After executing a selected action $a_t$, the ADS operates within an $OTP$, and then the operating environment is transitioned into a new state $s_{t+1}$. This experiment studies how different $OTP$ values impact \tool's performance.
As seen in Figure~\ref{fig:otp_analysis}, the optimal $OTP$ is $6$s with the highest number of $\#Collisons$ in the generated scenarios, and $TTC$ is also the lowest. Specifically, with $OTP = 6s$, the number of $\#Collisions$ is 11.8 which 64\% better than $OTP = 8s$ and nearly 10 times better than $OTP = 4$. Additionally, with $OTP=6$, the collisions occurred earlier than the other settings, from 11\%  to 36\% (lower TTC).

In practice, a shorter OTP (e.g., 4s) indicates a higher frequency of environment changes, preventing the RL agent from recognizing patterns or understanding the relationship between cause (applied actions) and effect (the transition of states). This hinders the RL agent's ability to learn how specific actions lead to dangerous situations over time, resulting in less effective scenario generation. Meanwhile, a long OTP (e.g., 8s) keeps the RL agent operating in a static environment for extended periods. This requires the RL agent to wait for a long time for feedback on its selected actions. This delay could make it difficult for the RL agent to connect actions with outcomes effectively. 
Therefore, it is essential to balance between stability and feedback frequency (e.g., $OTP = 6s$ in this experiment) to enable the RL agent to obtain the optimal results.

\subsubsection{$R_{col}$ Analysis}

\begin{figure}
    \centering
    \includegraphics[width=\linewidth]{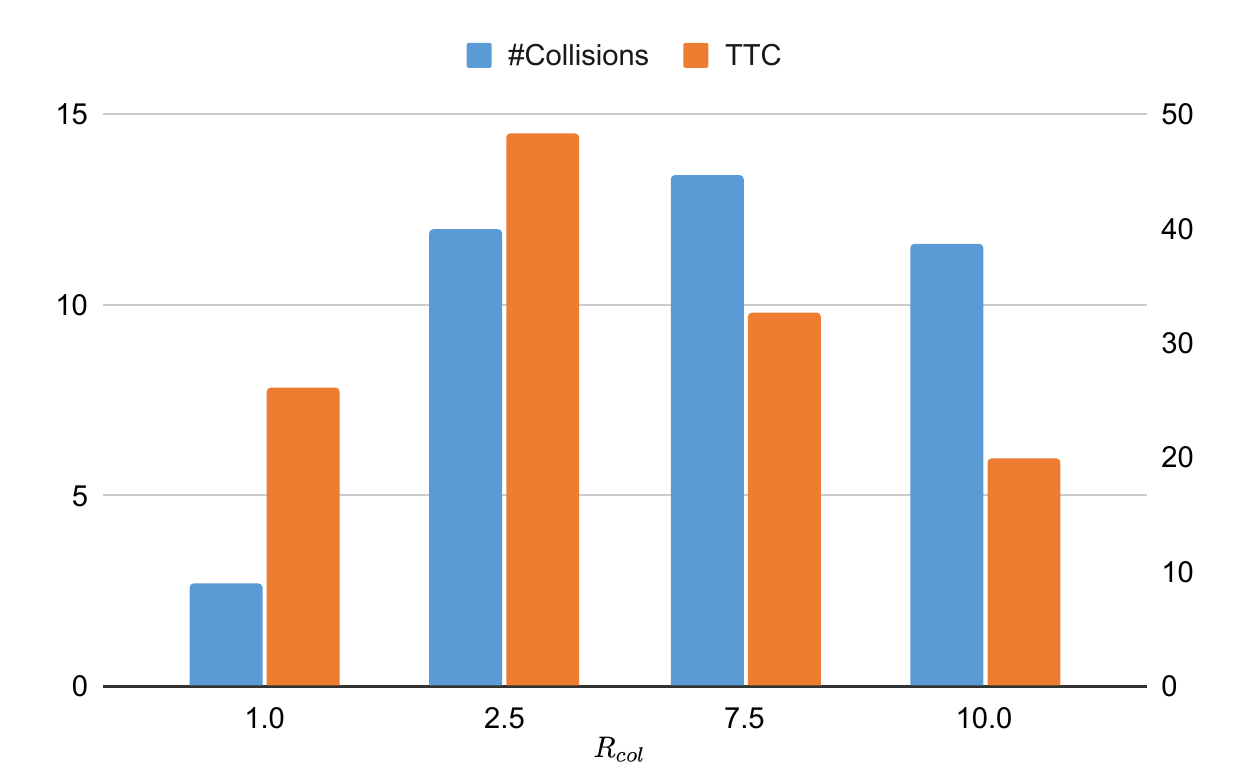}
    \caption{Impact of $R_{col}$ on \tool's performance (\textit{\#Collisions}: Left axis; \textit{TTC}: Right axis)}
    \label{fig:rcol_analysis}
\end{figure}

As seen in Figure~\ref{fig:rcol_analysis}, higher values of $R_{col}$, awarded to the RL agent when its selected action leads to an actual collision,  generally lead to the better performance of \tool. Specifically, at $R_{col} = 1$, \tool obtains only 2.7 collisions in the generated scenarios. At $R_{col} = 7.5$, 
\tool achieves the best performance with 13.4 collisions, outperforming the other settings by 12\% to 400\%. This improvement is expected because a higher $R_{col}$ means the RL agent receives a significantly greater reward for selecting effective actions.
This helps the RL agents better distinguish the actions that actually lead to collisions, thereby enabling a more effective generation of critical scenarios.

However, an overemphasis on rewarding several actions could also hinder the RL agent's learning capability. For instance, the number of collisions decreases by 14\% from 13.4 collisions at $R_{col} = 7.5$ to 11.6 collisions at $R_{col} = 10$. The reason is that a very high $R_{col}$ may cause the RL agent to overly prioritize certain actions over others, leading to more conservative choices. As a result, this limits the agent's flexibility in decision-making and reduces its ability to take necessary risks to explore other potentially optional actions.

\subsubsection{$\theta_\epsilon$ Analysis}

\begin{figure}
    \centering
    \includegraphics[width=\linewidth]{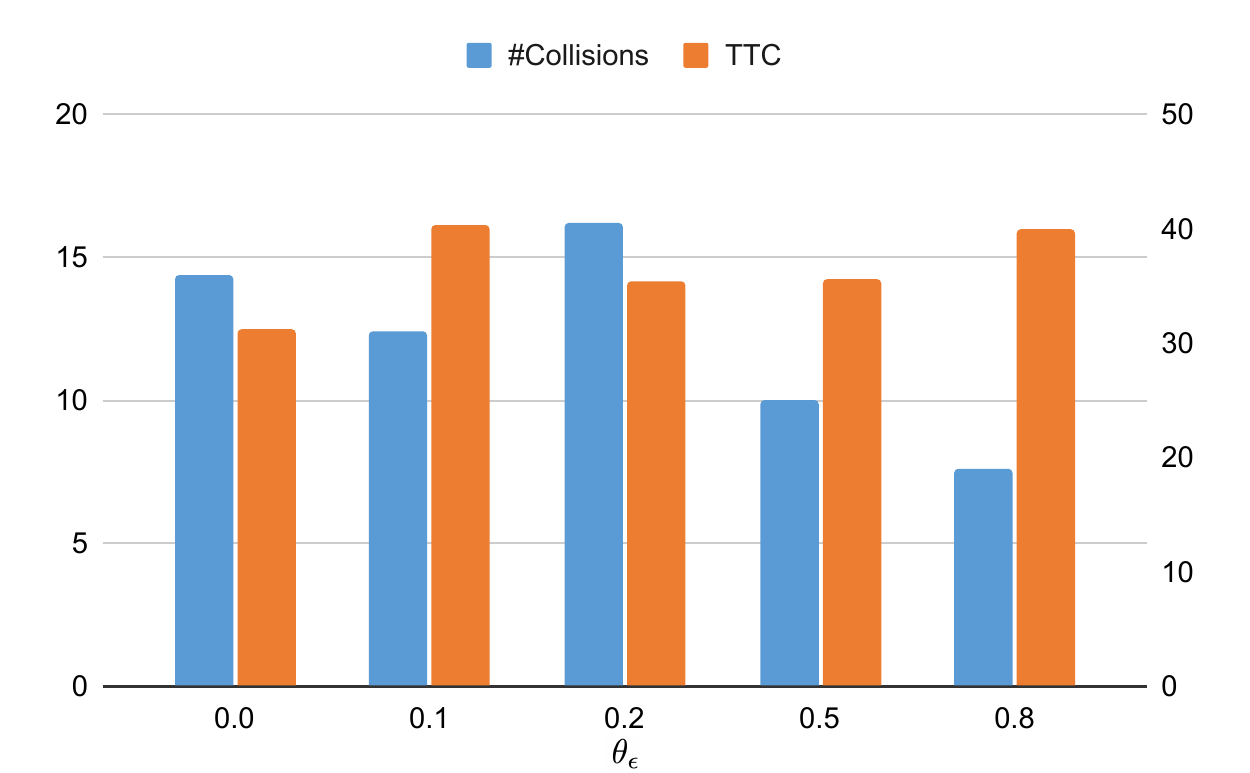}
     \caption{Impact of $\theta_\epsilon$ on \tool's performance (\textit{\#Collisions}: Left axis; \textit{TTC}: Right axis)}
    \label{fig:epsilon_analysis}
\end{figure}
 
To avoid premature convergence to local optima, \tool employs the $\epsilon$-greedy exploration strategy. Initially, $\epsilon$ is set to 1.0, encouraging full random exploration as the agent learns the environment dynamics. As training progresses, $\epsilon$ is gradually reduced to a threshold $\theta_\epsilon$ to balance exploration with exploitation. This experiment examines how different values of this threshold impact \tool's performance.

As seen in Figure~\ref{fig:epsilon_analysis}, 
a higher $\theta_\epsilon$ value results in fewer collisions.
Specifically, \textit{\#Collisions} decreases 2 times from 14.4 collisions when $\theta_\epsilon = 0.0$ to 7.6 collisions when $\theta_\epsilon = 0.8$.
The reason is that the higher the $\epsilon$ value, the more frequently the agent explores by selecting random actions. 
This could lead to many ineffective actions. 
Instead, when  $\epsilon$ is gradually reduced to a very small value, i.e., $\theta_\epsilon = 0.0$, 
the agent mainly exploits the learned behaviors, which could lead to more effective actions.

Interestingly, \tool reaches a peak of about 16 collisions with $\theta_\epsilon = 0.2$. In this setting, the agent follows its learned policy most of the time (80\%) but occasionally explores with random actions (20\%). This balance helps 
to encourage the agent to not only select collision-prone actions but also attempt slightly riskier or suboptimal behaviors that could also lead to collisions. This observation highlights the importance of carefully tuning exploration and exploitation to achieve optimal performance.

\begin{gtheorem}
\textbf{Answer to RQ3}: Each parameter significantly influences \tool's performance, making it crucial to thoughtfully configure them  to achieve optimal results.
\end{gtheorem}

\subsection{Answer RQ4: Sensitivity Analysis}

To evaluate \tool's sensitivity to different training environments, we measured its performance on a fixed testing road (Road 4)
under 3 training configurations: \textit{Cross-Road} Training (Roads 1, 2, and 3), \textit{Same-Road} Training (Road 4), and \textit{Comprehensive-Road} Training (all 4 roads). 

\begin{figure}
    \centering
    \includegraphics[width=\linewidth]{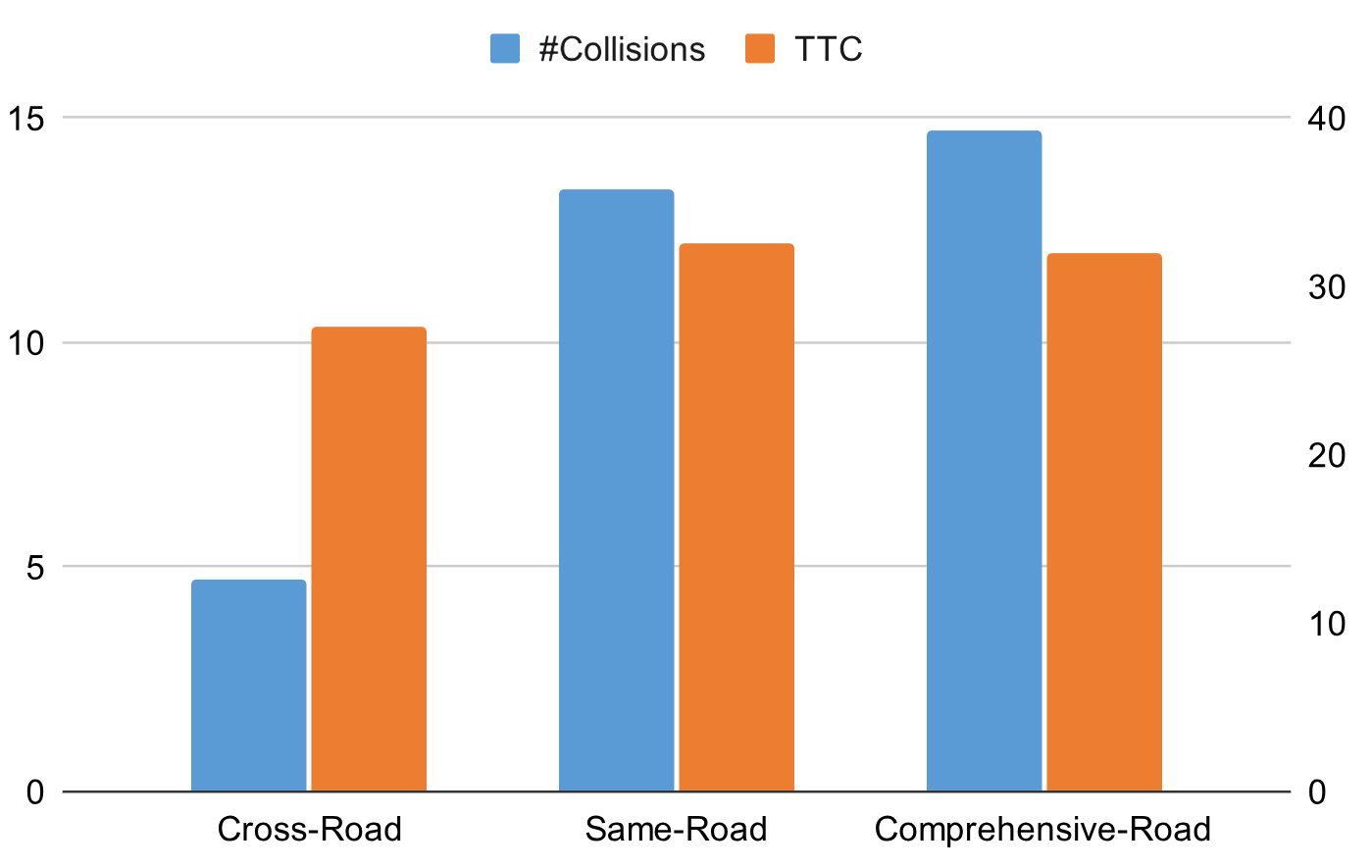}
    \caption{Impact of training settings on \tool's performance (\textit{\#Collisions}: Left axis; \textit{TTC}: Right axis)}
    \label{fig:RQ5-training}
\end{figure}

As shown in Figure~\ref{fig:RQ5-training}, \tool performs the least effectively in the \textit{Cross-Road} setting. While \tool demonstrates its ability to detect potentially risky situations quickly (short $TTC$), it fails to adapt the knowledge experienced on Roads 1, 2, and 3 to generate critical scenarios in a totally new road, Road 4. The lack of familiarity with the road-specific features results in \tool's less effective performance. 
The number of collisions in the scenarios generated in this setting is one-third that of the \textit{Same-Road} setting and 68\% lower than in the 
\textit{Comprehensive-Road}  setting.

In the \textit{Same-Road} setting, where the RL agent is trained and tested on the same road, the RL agent can gain detailed knowledge of Road 4's specific features. This familiarity allows \tool to adapt its experiences better to generate scenarios that more accurately reflect the actual dangers of the testing road. As a result, the generated scenarios are more challenging with 13.4 collisions, nearly three times as many as in the \textit{Cross-Road} setting.

The \textit{Comprehensive-Road} setting, which involves training on all four roads, results in the highest number of collisions  (i.e., 14.7 collisions).
 By combining exposure to both diverse and road-specific environments, \tool achieves a balance between generalization and specialization. This enables \tool to generalize across varied road layouts while retaining the specificity required for effectively generating high-risk scenarios on Road 4. Notably, the $TTC$ metric remains consistent across all training settings, indicating that scenario timing is relatively stable regardless of variations in training environments.

These findings reveal that while diverse training on unrelated roads enables \tool to detect potentially risky scenarios, training in the same environment significantly enhances its capacity to generate high-risk situations. The \textit{Comprehensive-Road} setting offers the optimal balance, combining the generalization benefits of diverse training with the relevance of specific exposure. A comprehensive training approach that includes diverse and environment-specific data is recommended for applications prioritizing the generation of high-risk scenarios. This balance maximizes \tool's ability to effectively challenge and evaluate ADSs across a range of realistic conditions.

\begin{gtheorem}
\textbf{Answer to RQ4}: The variety of training environments greatly affects \tool's ability to generate critical scenarios. Among the studied settings, the \textit{Comprehensive-Road} setting achieves the best results by balancing generalization and specialization, highlighting the need for both diverse and specific training data to maximize \tool's overall performance.
\end{gtheorem}

\subsection{Threats to Validity}
The main threats to the validity of our work consist of three parts: internal, construct, and external validity threats.

\textbf{Threats to internal validity} mainly lie in the correctness of our \tool's implementation. 
To address this, we rigorously reviewed our code and ensured its correctness through multiple verification rounds.
Additionally, we made the source code publicly available~\cite{website}, allowing other researchers to review and reproduce our experiments, which helps reduce potential errors.

\textbf{Threats to construct validity} are related to the rationality of the assessment metrics. To mitigate this threat, we selected the popular metrics such as $\#\textit{Collisions}$ and $\textit{TTC}$, which have been consistently used in prior research~\cite{mosat, av-fuzzer, deepcollision} to assess the effectiveness and efficiency of ADS testing approaches. To improve the accuracy of evaluation, we excluded subjective collisions, which the AV cannot avoid regardless of its capabilities. One potential threat may arise from the subjective assessment of what constitutes a collision. To minimize this threat, we carefully analyzed the generated scenarios and excluded unavoidable collisions.

\textbf{Threats to external validity} mainly lie in the roads and maps used in our experiments. This work conducted experiments on four roads where vehicles drive on the right-hand side of the road. Thus, the obtained results may not be generalized to other roads. In addition, we cannot claim that similar results would have been observed on the roads where vehicles must drive on the left-hand side. To migrate this threat, we selected popular maps widely used in related studies~\cite{mosat, av-fuzzer, deepcollision}. We also configured origins and destinations to construct diverse experimental road configurations. Another threat arises from the ADS and simulator in use. Our experiments were conducted on an industrial scale ADS, Apolo, and popular LGSVL simulator. Thus, the results may not be generalized to the other ADSs and simulators. In future work, we plan to evaluate the proposed approach on different ADSs and simulators across various maps.
\section{Related Work}
\label{sec:related_work}

Numerous testing approaches~\cite{kutila2018automotive, lim2018autonomous, wang2021can, li2020adaptive, arcaini2021targeting, zhang2022incremental, laurent2023parameter, gambi2019automatically, riccio2020model, corso2019adaptive, mosat, deepcollision, av-fuzzer, wu2024reality, lu2024diavio,ding2023survey, queiroz2024driver, birchler2024does} have been proposed to improve the safety and reliability of the ADS at both module-level and system-level. 

\textbf{Critical Scenario Generation.}
\tool is closely related to the approaches introduced to generate critical scenarios for testing ADS under different conditions. The simulation scenarios can be re-constructed from real-world dataset~\cite{huynh2019ac3r, dai2024sctrans, guo2024sovar, queiroz2024driver} or generated by identifying suitable environment configurations~\cite{mosat, av-fuzzer, deepcollision, lu2021search}. In the former direction~\cite{huynh2019ac3r, dai2024sctrans, guo2024sovar}, existing information, such as accident reports and videos, is analyzed to extract objects relevant to collisions. After that, corresponding objects, e.g., NPC vehicles or pedestrians, are generated and integrated into the simulation map accordingly. The generated scenarios can be utilized to test the ADS by replacing an NPC vehicle with an ego vehicle. 

Another direction focuses on environment configuration to generate adversarial scenarios. MOSAT~\cite{mosat} and AV-Fuzzer~\cite{av-fuzzer} generate testing scenarios by applying genetic search algorithms to find the maneuvers of the surrounding NPC vehicles that could lead the AV into a collision. \dc~\cite{deepcollision}  adopts RL to learn how to interact with the environment and create challenging scenarios. In another research, Hampen \etal~\cite{hempen2018model} employ a model-based method to generate test scenarios. They build a test model from the requirements, expand it with the states of the driving tests, and then generate complete driving scenarios to cover every possible state of the system. 

Aligned with these studies, \tool focuses on generating critical scenarios that maximize the revelation of weaknesses in the ADS. However, different from these studies, \tool takes into account not only the external factors but also the internal properties of the ADS components to better capture the collision causes. This enables the RL agents to find suitable values to configure the environment parameters more effectively. In addition, \tool considers both the challenging and realistic factors when generating collisions, making the generated scenarios more applicable in practice.

\textbf{System-level ADS testing.}
System-level testing focuses on testing the operation of the whole vehicle through collaborations between its modules. The system-level testing can be conducted in a simulator, on-road, or mixed reality.  

\textit{Simulation-based} testing is very popular in both industry and academia~\cite{tang2023survey, lou2022testing, mosat, av-fuzzer, abdessalem2018testing, gambi2019automatically, riccio2020model, corso2019adaptive, sato2021dirty, deepcollision, li2024scenarionet} due to its safety, cost-effectiveness, and the availability of powerful simulators. Search-based techniques~\cite{mosat, av-fuzzer, abdessalem2018testing, gambi2019automatically} are extensively used to search for parameter values that achieve a testing objective such as feature interaction failures~\cite{abdessalem2018testing}, dangerous scenarios~\cite{mosat, av-fuzzer}, or abnormal system behaviors~\cite{gambi2019automatically}. Moreover, model-based~\cite{riccio2020model}, stress testing~\cite{corso2019adaptive}, and adversarial attack~\cite{sato2021dirty} are also widely employed for testing ADS in the simulation environment.

Before deployment, ADS must be tested on \textit{long-distance public roads} to ensure its safety and reliability. Several AV manufacturers, such as Waymo and Tesla, have been testing their AVs on public roads in the US for years~\cite{zhao2019assessing, waymo2017, lou2022testing}. An ADS may be required to be tested from 50K to 100K kilometers on-road under different driving scenes, including highways, country roads, and urban areas. Despite the critical importance of on-road testing, only a small portion of ADS practitioners have done long-distance testing due to technical and financial constraints~\cite{lou2022testing}.

To balance between safety, reliability, effectiveness, and cost, \textit{mixed-reality} testing, which combines simulation-based and real-world testing, can be leveraged. 
To generate realistic scenarios for testing ADS in the simulators, multiple approaches~\cite{huynh2019ac3r, dai2024sctrans, guo2024sovar} have collected real-world data and employed them in scenario generation. Additionally, several real physical components can be integrated into the testing loop, such as ECU hardware~\cite{gao2020hardware}, real vehicles~\cite{li2021real}, or pedestrian dummies~\cite{szalay2023critical}.


\tool contributes to system-level testing by focusing on the generation of critical scenarios that test the ADS in a simulator, ensuring the scenarios are challenging and reflective of real-world dynamics.

\textbf{Module-level ADS testing.}
An ADS typically comprises various sensors and modules such as localization, perception, planning, and control~\cite{first-look, tang2023survey}. To guarantee the quality and correctness of each module, multiple testing techniques have been introduced~\cite{kutila2018automotive, wang2021can, kumar2020black, eykholt2018robust, song2018physical, arcaini2021targeting, calo2020generating, laurent2023parameter, li2024chatgpt}. 
For \textit{hardware equipment}, Kutila~\etal~\cite{kutila2016automotive,kutila2018automotive} test the capabilities of sensors under harsh weather conditions such as rain and fog.  Besides, several works perform deliberate attacks such as jamming attack~\cite{lim2018autonomous} or spoofing attack~\cite{wang2021can} to test the reliability of these devices. 

Among software modules, the \textit{perception module} has gained the most attention from researchers~\cite{tang2023survey}. This module employs various ML/DL models to detect nearby objects and predict their movements, playing a crucial role in the ADS's operation. However, it is vulnerable to safety and security threats. 
\textit{Adversarial attack} is the main approach for testing the ML/DL models in the perception module. For example, Zhao \etal~\cite{wang2021can} generate perturbations to make object detectors unable to recognize objects and make incorrect recognition. Similarly, Li \etal~\cite{li2020adaptive} and Kumar \etal~\cite{kumar2020black} introduce approaches to attack traffic sign recognition models. While the adversarial attack methods effectively expose weaknesses in ML/DL models, they often overlook the realism of the perturbations and noisy physical environments. To address this, Eykholt \etal~\cite{eykholt2018robust} and Song \etal~\cite{song2018physical} propose Robust Physical Perturbation methods which perturb only the object (e.g., a road sign) without altering its surrounding environment. The experimental results demonstrate that their generated adversarial examples are robust under different physical environment conditions.

\section{Conclusion}
\label{sec:conclusion}

This paper proposes \tool, an RL-based approach for generating realistic critical scenarios that holistically account for both the internal states of the ADS and the external states of the  surrounding factors. 
\tool introduces a diverse set of actions that allows the RL agent to systematically configure both \textit{environmental conditions} and \textit{traffic participants}.
In addition, \textit{to ensure the realism of the generated scenarios, we incorporate several heuristic constraints}, ensuring the scenarios remain plausible and relevant for testing.
We evaluate \tool on two popular simulation maps with four road configurations, demonstrating its ability to outperform state-of-the-art approaches like \dc by generating 30\% to 115\% more collision scenarios. Furthermore, compared to Random Search, \tool achieves up to 275\% better performance. These results highlight the effectiveness of \tool in enhancing the safety testing of AVs through realistic, comprehensive scenario generation.

\bibliographystyle{IEEEtran}

\bibliography{references}

\end{document}